\documentclass[final]{cvpr}

\usepackage{times}
\usepackage{epsfig}
\usepackage{graphicx}
\usepackage{amsmath}
\usepackage{amssymb}
\usepackage{graphicx}
\usepackage{multirow}
\usepackage{amsmath,amssymb,amsfonts}
\usepackage{subcaption}
\newtheorem{theorem}{Theorem}
\newtheorem{corollary}{Corollary}[theorem]
\usepackage{algorithm, algpseudocode}
\usepackage{color}

\usepackage[pagebackref=true,breaklinks=true,colorlinks,bookmarks=false]{hyperref}

\def\cvprPaperID{5546} 
\def\confYear{CVPR 2021}

\newcommand{\prl}{MCOSS}
\newcommand{\oursubmodular}{SubMCOSS}
\newcommand{\newformulation}{ThreshMCOSS}

\begin{document}

\title{Convex Online Video Frame Subset Selection using Multiple Criteria for Data Efficient Autonomous Driving}

\author{
Soumi Das, Harikrishna Patibandla\\
IIT Kharagpur\\
Kharagpur, India\\
{\tt\small soumi\_das@iitkgp.ac.in}\\ 
{\tt\small krishnahari051197@gmail.com}
\and
Suparna Bhattacharya, Kshounis Bera\\
Hewlett Packard Enterprise India\\
Bangalore, India\\
{\tt\small suparna.bhattacharya@hpe.com,}\\
 {\tt\small kshounis.bera@hpe.com}
\and
Niloy Ganguly, Sourangshu Bhattacharya \\
IIT Kharagpur\\
Kharagpur, India\\
{\tt\small niloy@cse.iitkgp.ac.in,}\\
{\tt\small sourangshu@cse.iitkgp.ac.in}
}

\maketitle

\begin{abstract}
   Training vision-based Urban Autonomous driving models is a challenging problem, which is highly researched in recent times.
   Training such models is a data-intensive task requiring storage and processing of vast volumes of (possibly redundant) driving video data.
  In this paper, we study the problem of developing data-efficient autonomous driving systems. In this context, we study the problem of multi-criteria online video frame subset selection.
  We study convex optimization based solutions and show that they are unable to provide solution with high weightage to loss of selected video frames. We design a novel convex optimization based multi-criteria online subset selection algorithm which uses a thresholded concave function of selection variables. We also propose and study submodular optimization based algorithm. 
  Extensive experiments using driving simulator CARLA shows that we are able drop 80\% of the frames, while succeeding to complete 100\% of the episodes w.r.t. the model trained on 100\% data, in the most difficult task of taking turns. 
  This results in training time of less than 30\% compared to training on the whole dataset. We also perform detailed experiments on prediction performances of various affordances used by the Conditional Affordance Learning (CAL) model, and show that our subset selection improves performance on the crucial affordance \textit{``Relative Angle''} during turns.
\end{abstract}

\section{Introduction}

Many A.I. applications, including autonomous driving systems \cite{sauer2018conditional,dosovitskiy2017carla}, process large amount of video data for training complex deep learning models. However, much of the input video contains redundant information from the task point of view. For example, in case of autonomous driving, training using many frames on straight road sections may not be necessary; while one may need a lot of frames in the turns for training. In this paper, we study the problem of selecting video frame subsets, which are most informative as training data for one or more such tasks. 
We are also interested in developing a  selection algorithm that can be deployed on edge devices, thus minimizing edge to core transfer.

Autonomous driving models typically involve multiple prediction tasks, e.g. prediction of 6 affordances in conditional affordance learning (CAL) model \cite{sauer2018conditional}. The importance of prediction accuracy of a particular task towards the end objective  is a complex function of the situation. For example, we find that error in prediction of relative angle during the turns is critical towards achieving higher episode completion, since the vehicle is able to easily recover from such errors during straight stretches. Hence, we are interested in designing a video frame subset selection algorithm which is able to incorporate multiple criteria arising from the different tasks, situations (e.g. turns, presence of pedestrians, etc) and existing model properties (e.g. prediction loss).

The problem of video frame subset selection has been studied in various contexts, including video summarization \cite{elhamifar2017online,mahasseni2017unsupervised}, video recognition\cite{Wu_2019_CVPR}, video fast forwarding \cite{lan2018ffnet}, 
etc. Broadly, the techniques can be classified into two classes: (1) deep reinforcement learning based methods, which learn a frame skipping network, with reward for better performance \cite{lan2018ffnet,Wu_2019_ICCV}, or more confidence \cite{Wu_2019_CVPR} on the end task; and (2) external criteria based methods which optimize a global criterion between selected frames and whole video, e.g. perceptual similarity \cite{elhamifar2017online}, manifold spanning \cite{joneidi2020select}, etc. For our setting, the first class of methods are too expensive, since they require multiple evaluations of the end task objective, which in our case is the fraction of test episodes completed.

In this paper, we build on the online subset selection framework \cite{elhamifar2017online}, where at each step, an existing set of selected frames is supplemented by the most relevant frames from an incoming set, based on the total dissimilarity between selected frames and incoming frames. This framework was extended to incorporate pointwise loss of the selected frames in a composite criteria \cite{das2020multi}. These techniques were applied to video summarization \cite{elhamifar2017online} and semantic segmention \cite{das2020multi}. To the best of our knowledge, these techniques have not been used in the autonomous driving setting. Moreover, through a rigorous analysis, we show that, the additive incorporation of pointwise loss criteria in \cite{das2020multi} suffers from selection of fewer frames from incoming set as we provide higher weightage to the pointwise criteria. This is due to the fact that additive incorporation suffers from multiple counting of loss of selected points thus leading to selection of fewer frames. We propose a novel thresholded formulation (\newformulation) of incorporation of pointwise criteria, which leads to a convex selection objective leading to efficient algorithm.

We also propose a natural set-function based criteria for incorporation of pointwise multiple criteria, which is also shown to be a sub-modular function, thus opening up the possiblity of using approximate submodular optimization algorithms. In an thorough empirical study, we show that both on synthetic data, as well as data from selection of video frames for autonomous driving, the convex relaxation based algorithm performs better than state of the approximations algorithms for submodular maximization \cite{buchbinder2014submodular,buchbinder2015tight}. 

We perform the autonomous driving experiments by training on a dataset of more than 100,000 video frames collected from our testbed of CARLA simulator \cite{dosovitskiy2017carla}, and testing on 40 episodes of from 4 different settings, with more than 6000 video frames. Through extensive experimentation, we find that bucketwise relative angle loss as a good criteria along with total multi-task loss and SIFT similarity between pairs of frames provides an ideal selection criteria. Our experiments using the CAL model \cite{sauer2018conditional}, show that we can achieve a 100:20 compression (selecting 1 in 5 frames) which does not lead to any loss in episode completion, even though the accuracies on individual tasks decrease marginally.

\subsection{Related Work}

We describe two broad classes of relevant prior works: (a) Learning model for self driving cars, and (b) Subset selection.
The autonomous driving literature has witnessed many different kinds of learning approaches: (1) modular pipeline or mediated perception \cite{dosovitskiy2017carla} \cite{urmson2008autonomous} \cite{buehler2009darpa} \cite{thrun2006stanley} \cite{ziegler2014making}, (2) imitation learning \cite{pomerleau1989alvinn} \cite{bojarski2016end} \cite{codevilla2018end} \cite{bojarski2017explaining} \cite{xu2017end}, and reinforcement learning \cite{dosovitskiy2017carla} \cite{liang2018cirl} \cite{prakash2020exploring}.
Deep Imitation Learning based method 
Conditional Affordance Learning (CAL) \cite{sauer2018conditional} learns models for multiple affordances in a multi-task setting, using driving video data collected from the driving simulator CARLA \cite{dosovitskiy2017carla}. It claims better episode completion over other imitation learning based approaches e.g. CARLA \cite{dosovitskiy2017carla} or conditional imitation learning \cite{codevilla2018end}. Hence, in this work we demonstrate effectiveness of our frame subset selection algorithms using CAL as the learning model with driving simulator CARLA \cite{dosovitskiy2017carla}. However, our approach is complementary to recent reinforcement learning based approaches e.g. \cite{prakash2020exploring}, as discussed in section \ref{sec:method}.
\noindent \textbf{Video frame subset selection:}\\
Recent online subset selection approaches can be divided into two broad classes (1) deep learning based, and (2) based on optimization of some input criteria.
The first class of techniques \cite{lan2018ffnet}, \cite{Wu_2019_ICCV} \cite{Wu_2019_CVPR} depend on selection networks added to the pipeline of existing tasks and are trained jointly. They are able to learn complex selection criteria through deep models, but do not come with any stated selection criteria, w.r.t. which the selection is optimal.
These methods are jointly trained with the end objective, e.g. video recognition for \cite{Wu_2019_ICCV} and \cite{Wu_2019_CVPR}, and are typically too expensive to be deployed on edge devices. 

The second class of techniques, which are also closest to us, selects the data points based on different defined criteria. These criteria include reconstruction error \cite{chen2012surveillance}, linear dependency \cite{joneidi2020select}, perceptual similarity \cite{elhamifar2017online} or criteria based on end tasks such as distinctiveness and uncertainty \cite{huang2018cost} \cite{das2020multi}. 
\cite{joneidi2020select} recently proposed an online approach based on linear dependence criteria. However for the current problem, we build on the pairwise criteria based approach proposed in \cite{elhamifar2017online}, and extended to include multiple pairwise and pointwise criteria \cite{das2020multi}. While these approaches were used for the problems of Video summrization \cite{elhamifar2017online} and semantic segmentation \cite{das2020multi}, their setting provides flexibility of incorporating multiple different types of criteria, which is relevant to our application. We provide detailed comparison with these approaches.

\section{Data Efficient Autonomous Driving}
\label{sec:method}
\newcommand{\cA}{\mathcal{A}}
\newcommand{\cG}{\mathcal{G}}
\newcommand{\bz}{\mathbf{z}}

\newtheorem{remark}{Remark}

In this section, we describe the problem of data efficient training of autonomous driving models, with the core idea being selection of relevant video segments in an online setting. We formulate this problem as an online subset selection problem (OSS) for selecting subsets of video frames, given input signals from existing selected video frames, and the trained model, which throw up multiple input criteria - both for pairs of input frames and for single input frame. Section \ref{sec:setup} describes the setup of data efficient autonomous driving and its connection to multi-criteria OSS problem formulation. Section \ref{sec:analysis} points out a drawback with existing multi-criteria OSS formulation. Sections \ref{sec:submodular} and \ref{sec:newformulation} describe our new formulations for the multi-criteria OSS problem.

\subsection{Problem setup and OSS}
\label{sec:setup}

Training of vision based autonomous driving models \cite{sauer2018conditional, prakash2020exploring} requires processing of large amounts of annotated video data. 
In many cases, the videos are collected in episodes over a period of time, leading to processing and training of models in batches which are ordered in time.
Hence, the batchwise OSS scheme discussed in \cite{elhamifar2017online}, is an ideal setting for selection of video frames in this context.
 We denote a complete dataset as $\mathcal{D} = \{ (x_i,y_i),\ i=1,\dots,n \}$ where $n$ is the total number of datapoints (annotated video frames) in the dataset, $x_i$ are the features extracted from video frames, and labels $y_i$ corresponding to various learning tasks, e.g. affordances \cite{sauer2018conditional}. 
Let $X_t = \{ (x_i,y_i),\ i=1,\dots, m\} ,\ t = 1,\dots,T$ denote the $t^{th}$ batch of episodes collected, where $m$ denotes the number frames in a batch \footnote{Equal batch size is for simplicity of exposition, not a requirement}. Hence $mT = n$.  We also define the cumulative sets $C_t = \cup_{i=1}^t X_i$ denoting all  data collected till batch $t$. We are interested in constructing representative set $R_t \subseteq C_t$, which consists of a representative set of frames till batch $t$. The intention here is that an autonomous driving model $M_t$ trained on the cummulative set $C_t$, should perform similarly to another model $M'_t$ trained on the representative set of frames $R_t$ in terms of an end performance metric, e.g. the episode completion metric used in \cite{sauer2018conditional}. Furthermore, the  size of $R_t$ should be small so that $R_t$ uses lower storage space and communication bandwidth, and training of $M'_t$ potentially takes lower time. An algorithm for selection of $R_{t+1}$ from $X_{t+1}$, given $C_{t}$,  and $M'_{t}$ thus constitutes a \textit{data efficient scheme} for training of autonomous driving models, since we are only storing and processing $R_t$'s. Note that, this scheme can also be used in the reinforcement learning schemes for improvement of driving policies such as the one described in \cite{prakash2020exploring}, where $X_t$ can be taken from the replay buffer at iteration $t$.


For the OSS formulation, we focus on an input batch of episodes $X_{t+1}$. The selection algorithm uses two input sets of frames $R_t$ and $X_{t+1}$, here referred to as the old set (superscript $o$) and new set (superscript $n$) respectively, following notation used in \cite{elhamifar2017online}. Let $d^o_{ij}$ denote a dissimilarity measure between new frame $i$ (from $X_{t+1}$) and old frame $j$ (from $R_t$), and $d^n_{ij}$ denote the dissimilarity between new frames $i$ and $j$ (both from $X_{t+1}$). The OSS formulation minimizes the a composite criteria with two parts: (1) total dissimilarity between the "representative frames" (either from new or old set) and the incoming frame it represents, and (2) number of representative frames from the new set.
Let $z_{ij}^o, z_{ij}^n $ be the relaxed binary assignment variables ($Z_{ij}\in [0,1])$, where $z_{ij}^o=1 $ denotes that the representative of $i^{th}$ new example ( $(x_i,y_i)\in X_{t+1}$) is $j^{th}$ old example ($(x_j,y_j)\in R_t$), and $z_{ij}^n = 1$ denotes that representative of $i^{th}$ new example ($(x_i,y_i)\in X_{t+1}$) is the $j^{th}$ new example ($(x_j,y_j)\in X_{t+1}$). Otherwise, $z_{ij} = 0$. Hence any solution for optimal representative allocation should satisfy the constraint: $\sum_{j=1}^{|R_t|} z_{i,j}^o + \sum_{j=1}^m z_{i,j}^n = 1$, asserting that every frame  $i\in X_{t+1}$ has exactly one representative. The objective function can be written as:
\small
\begin{align}
    L(z^o_{ij},z^n_{ij}) = \sum_{i=1}^m \sum_{j=1}^{|R_t|} z_{ij}^{o} d^o_{ij} + \sum_{i,j=1}^m z_{ij}^n d^n_{ij} + \lambda \sum_{j=1}^m \| [z_{1,j}^n \dots z_{m,j}^n ] \|_p \nonumber
\end{align}
\normalsize
Das et al. \cite{das2020multi} has incorporated both pairwise scores (e.g. distance $d_{ij}$ between pairs of frames $i,j$) and pointwise scores (e.g. negative loss $-L_i$ for the frame $i$). The modified \textit{cumulative dissimilarity} function $Q_{ij}$ is a weighted sum of $d_{ij}$ and $L_j$ - the loss incurred by the representative point. Thus $Q_{ij} = \rho d_{ij} - (1-\rho) L_j$.
Let $L^n_i$ denote the pointwise attribute (here loss value) for datapoint $i$ in $X_{t+1}$ and analogously for $L^o_i$ (denoting loss for datapoint $i$ in $R_t$). The final formulation is:

 \small
\begin{align}
\label{eqn:oldsiftloss}
    &\min_{z^o_{ij},z^n_{ij}} \sum_{i=1}^m \sum_{j=1}^{|R_t|} z_{ij}^{o} Q^o_{ij} + \sum_{i,j=1}^m z_{ij}^n Q^n_{ij} + \lambda \sum_{j=1}^m \| [z_{1,j}^n \dots z_{m,j}^n ] \|_p \nonumber \\ 
    &s.t.\sum_{j=1}^{|R_t|} z_{i,j}^o + \sum_{j=1}^m z_{i,j}^n = 1, \ \forall i\in X_{t+1} \nonumber \\
    & z_{i,j}^n, z_{i,j}^o \in [0,1], \ \forall i,j
\end{align}
\normalsize
where $Q^n_{ij} = \rho d^n_{ij} - (1-\rho) L^n_j$ and $Q^o_{ij} = \rho d^o_{ij} - (1-\rho) L^o_j$.
This is a convex optimization problem which can be solved efficiently for moderate sizes of sets $X_{t+1}$ and $R_{t}$ using off the shelve solvers, e.g. CVXPY \cite{diamond2016cvxpy}. We call this formulation \textit{multi-critria OSS} (\prl).

\subsection{Analysis of Multi-Criteria OSS}
\label{sec:analysis}

While applying \prl\ to our problem, we noticed that as we give higher weightage to the pointwise component by choosing lower value of $\rho$, the number of selected points decreases. From an application point of view, this allows the pointwise score to have a limited impact on the set of points selected. This might be tolerable in certain applications, e.g. semantic segmentation where the perceptual dissimilarity measure contains sufficient information for frame subset selection. However for the application of autonomous driving, we find that task-wise and situation-wise losses have much more impact on the quality of frames selected.

To understand the mechanism through which this problem arises, we observe that using only pointwise metric yields a maximum of one representative for all images belonging to incoming set $X_{t+1}$.
This is the setting when $\rho$ = 0. We have  $Q_{ij}^o$ = $- (1-\rho) L_j^o$ and $Q_{ij}^n$ = $- (1-\rho) L_j^n$, both of them are constant across $i$. 
The representative $j$ of any instance $i\in X_{t+1}$ $(i = 1, ..., m)$ will be from $X_{t+1}$ if $L_j^n > L_{j'}^ \forall j'\in X_{t+1}$ in which case only one point will be selected (see Corollary \ref{coro1}). Otherwise the representative will be from $R_{t}$, in which case no points are selected.



While the above intuitions are motivated for special case of $\rho=0$, the ideas also apply to more general values of $0<\rho\leq 1$. We further illustrate this by characterising the solution of formulation \ref{eqn:oldsiftloss} in the following theorem. 

\begin{theorem}
\label{thm:representative-condition}
Let $z_{ij}^{o}$ and $z_{ij}^{n}$ be the optimal solution for formulation \ref{eqn:oldsiftloss}.
A new frame $j\in X_{t+1}$ is selected as a representative frame for at least one incoming frame $i\in X_{t+1}$, i.e. $z_{ij}^n = 1$, only if the following conditions hold:
\begin{itemize}
\item For some incoming frame  $i \in X_{t+1}$ , $Q^n_{ij} < Q^n_{ij'}$, for all $j'\in X_{t+1}$ and $j' \neq j$
\item For some incoming frame  $i \in X_{t+1}$, $ Q^n_{ij} < \frac{ \sum_{i'=1}^m z_{i',k}^o Q^o_{i'k} + \lambda \| [z_{1,j}^n \dots z_{m,j}^n ] \|_p }{\| \bz_{j}^n \|_1} $
\end{itemize}
where $k = argmin_j \sum_{i=1}^m z_{i,j}^o Q_{i,j}^o$, and $\| \bz_{j}^n \|_1 = \sum_{i'=1}^m z_{i'j}^n $
\end{theorem}

Due to space constraints, we provide the formal proof in the supplementary material. Note that the first condition states that there is at least one frame $i$ in the incoming set whose cumulative dissimilarity $Q^n_{ij}$ is lower than all other points. The second condition signifies that cumulative dissimilarity $Q^n_{ij}$ between a representative $j$ and the point it is representing $i$ is lower than minimal contribution from a potential representative $k$ from existing set of selected examples $k\in R_{t}$. Next, we provide two corollaries to illustrate our point. Corollary \ref{coro1} illustrates the conditions in theorem \ref{thm:representative-condition} for the special case of $\rho=0$. Since, the dependence on $i$ is removed, it is easy to see that at most one $j\in X_{t+1}$ will satisfy the condition.


\begin{corollary}
\label{coro1}
Let $z_{ij}^{o}$ and $z_{ij}^{n}$ be the optimal solution for formulation \ref{eqn:oldsiftloss}.
A new frame $j\in X_{t+1}$ is selected as a representative frame for at least one incoming frame $i\in X_{t+1}$, i.e. $z_{ij}^n = 1$, only if the following conditions hold:
\begin{itemize}
\item $L_j^n > L_{j'}^n$ for all $j'\in X_{t+1}$ and  $j' \neq j$
\item $L^n_j > \frac{ \sum_{i=1}^m z_{i,k}^o L_k^o - \lambda \| [z_{1,j}^n \dots z_{m,j}^n ] \|_p }{\| \bz_{j}^n \|_1 } $
\end{itemize}
where $k = argmin_j \sum_{i=1}^m z_{i,j}^o Q_{i,j}^o$, and $\| \bz_{j}^n \|_1 = \sum_{i'=1}^m z_{i'j}^n $
\end{corollary}

\begin{corollary}
\label{coro12}
Let $\Delta_d(i,j) = \|\bz_j^n\|_1 d_{ij}^n - \sum_{i'=1}^{m} z_{i'k}^o d_{i'k}^o$ and $\Delta_L(j) = \|\bz_j^n\|_1 L_{j}^n - \sum_{i'=1}^{m} z_{i'k}^o L_{k}^o$. If $\Delta_d(i,j) < - \Delta_L(j)$ for all $z_{ij}^n,z_{ij}^o$, and for $\rho=0$, $j\in X_{t+1}$ is not a representative frame, then for some $\rho \geq 0$, theorem \ref{thm:representative-condition} will not be satisfied by any pair $i,j\in X_{t+1}$.
\end{corollary}

Corollary \ref{coro12} states that if a frame $j\in X_{t+1}$ is not a representative, and satisfies the conditions on $\Delta_d(i,j)$ and $\Delta_L(j)$, then it will stop being a representative for some value of $\rho \geq 0$. By rearranging the terms in second condition of theorem \ref{thm:representative-condition}, we get: $\rho \Delta_d(i,j) - (1-\rho) \Delta_L(j) \leq \lambda\frac{\|\bz^n_j\|_p}{\|\bz^n_j\|_p}$. For $p=1$ the RHS is constant, but LHS decreases with $\rho$. Hence the second condition of theorem \ref{thm:representative-condition} is not satisfied by any $i\in X_{t+1}$ for the given candidate representative frame $j\in X_{t+1}$. These results motivate us to look for better formulations of multi-criteria OSS problem.



\subsection{Submodular Multi-Criteria OSS}
\label{sec:submodular}

In this section, we describe an algorithm for multi-criteria OSS problem based on submodular optimization. The problem can be posed as a set function incorporating both pairwise and pointwise attributes and can be solved using submodular optimisation. The natural criteria used for selection is the pre-defined modified \textit{cumulative dissimilarity} function $Q_{ij}$.

For every set $S$, the set function $f(S)$ can be defined as:
\small
\begin{equation}
    f(S) = \sum_{i \in X} min \{ min_{j \in R} Q_{ij} , min_{j \in S} Q_{ij} \} 
\label{eq:submod}
\end{equation}
\normalsize
Here, the problem is solved by selecting a representative $j$ which contributes the least dissimilarity value $Q_{ij}$ to the incoming instances $i \in X$. By definition, we can say
\begin{remark}
-f(S) is submodular.
\end{remark}
For proof, see Supplementary. We can thus pose it as a submodular maximisation problem by solving the problem $min_{S \subseteq X} f(S)$. We call this formulation \textit{submodular multi-criteria OSS} (SubMCOSS). 
We define a greedy submodular maximisation approach for solving the optimisation problem in Algorithm \ref{algo:submodular}. The algorithm is a randomised greedy algorithm that examines the dataset $k$ times to select the representatives for the incoming data. Next, we show a thresholded convex approach for solving the multi-criteria OSS problem.
\small
\begin{algorithm}[tb]
 \caption{: Submodular Multi-Criteria OSS}
 \label{algo:submodular}
 \begin{algorithmic}[1]
 
 \State \textbf{Input:}
 \State \hspace{2mm}$S_0$: Initial representative set
 \State \hspace{2mm}$X$: Incoming Set of Instances
 \State \hspace{2mm}$k$: Subset cardinality
 \State \hspace{2mm}$f(S)$: Function to be minimised
 \State \hspace{2mm}$S_0 = \phi$
 \State \textbf{Process:}
 \hspace{2mm}\For{$i$ = 1,2, $\ldots$, $k$}
 \hspace{2mm}\For{each x $\in$ X \textbackslash $S_{i-1}$}
 \State  $fv_x \leftarrow f(S_{i-1} \cup {x})$
  \EndFor
 \State  Let $M_i \in X$ \textbackslash $S_{i-1}$ be subset of top $k$ elements maximising $\sum_{m \in M_i}fv_m$
 \State  Let $u_i$ be randomly sampled from $M_i$ 
 \State $S_i \leftarrow S_{i-1}\cup {u_i}$
 \EndFor
 \State \textbf{Output:}
 \State $S_k$ : Subset of size $k$
 \end{algorithmic}
\end{algorithm}
\normalsize

\subsection{Thresholded Convex multi-criteria OSS}
\label{sec:newformulation}

\oursubmodular\ , described in previous section uses the natural formulation of weighted linear aggregation of pointwise and pairwise loss function. However, the algorithm for submodular optimization is a randomized approximation algorithm, and also computationally expensive due to multiple sampling runs required for a good optimal subset (see our results). In this section, we describe a novel convex formulation multi-criteria OSS which alleviates the problems of \prl\ (equation \ref{eqn:oldsiftloss}) as well as \oursubmodular\ (Algorithm \ref{algo:submodular}).

The key observation which helps us in designing the novel algorithm is that in \prl\ (equation \ref{eqn:oldsiftloss}) it is possible for a frame $j\in X_{t+1}$  to contribute $ - m (1-\rho) L_j^n $ by becoming a representative for every point $i\in X_{t+1}$ (see that the terms involving pointwise loss add up to $-  (1-\rho) (\sum_{i=1}^{m} z^n_{ij} ) L_j^n$).
However, in reality it only adds one data point to the training set with the pointwise score of $- L_j^n$. This problem is alleviated by using a coefficient of $L_j^n$ which is an indicator of whether $j$ is a representative point or not, rather than $(\sum_{i=1}^{m} z^n_{ij}$ which counts the number of points represented by $j$. This is achieved by using a concave function $S_j$ of $z_{ij}$: $S_j$ = $\frac{1}{\epsilon} \min (\epsilon , \sum_{i=1}^m z_{ij})$ where $\epsilon$ is an input parameter. The modified objective function becomes: $\cG(z_{ij}^o, z_{ij}^n) = \rho ( \sum_{i=1}^m \sum_{j=1}^{|R_t|} z_{ij}^{o} d^o_{ij}(t) + \sum_{i,j=1}^m z_{ij}^n d^n_{ij}(t) )  - (1-\rho) (\sum_{j=1}^{|R_t|} S_j^o * L_j^o  + \sum_{j=1}^m S_j^n * L_j^n )$, where, $S^o_j$ = $\frac{1}{\epsilon} \min (\epsilon , \sum_{i=1}^m z_{ij}^o)$ , $S^n_j$ = $\frac{1}{\epsilon} \min (\epsilon , \sum_{i=1}^m z_{ij}^n)$.
Note that $\cG$ is a convex function of $z_{ij}^o, z_{ij}^n$ since $S$ is a concave function. 
Also note that , each potential representative $j\in X_{t+1}$ can contribute a maximum of its own pointwise score $L_j$, since $S_j$ can take a maximum value of $1$. 

Another drawback of \prl\ (equation \ref{eqn:oldsiftloss}) is that compression ratio has no direct relation with the parameter $\lambda$. 
We use a constraint based cardinality criteria in order to have more precise control over the number of representative selected. The user provided parameter \textit{frac} specifies and upper bound over the fraction of incoming frames to be selected as representatives. 
Overcoming these drawbacks our final convex optimisation based multi-criteria OSS problem formulation is:

\small
\begin{align}
\label{eqn:newsiftloss}
    \min_{z^o_{ij},z^n_{ij}} & \cG(z_{ij}^o,z_{ij}^n) \\
    s.t. & \sum_{j=1}^{|R_t|} z_{i,j}^o + \sum_{j=1}^m z_{i,j}^n = 1 \nonumber \\
    &z_{i,j}^n, z_{i,j}^o \in [0,1] \nonumber \\ 
    &\sum_{j=1}^{m}\| [z_{1,j}^n \dots z_{m,j}^n ] \|_p \leq frac * m
    \nonumber
\end{align}
\normalsize

This can be efficiently solved using any convex solver, e.g. CVXPY \cite{diamond2016cvxpy}. We call this formulation \textit{thresholded multi-criteria OSS} (\newformulation).
$\epsilon$ is a user input which is designed to be the maximum value taken by the variable $\sum_i z_{ij}$, when none of the $z_{ij}$ denote a representative relation to be true. In an ideal situation (when we achieve a $\{0,1\}$ solution to $z_{ij}$), any positive value for $\epsilon$ is sufficient. In practise, we set $\epsilon$ to a value less than $1$, e.g. $\epsilon = 0.9$. 
Next, we experimentally demonstrate the utility of our method.

%

\section{Experiments}

In this section, we provide our experimental setup by describing the dataset and the subset selection methods. We then show a comparison of approximate online subset selection methods over synthetic data and collected driving data from the CARLA simulator \cite{dosovitskiy2017carla}. Next, we compare the performance of various subset selection methods in the autonomous driving application from two aspects: episode completion and affordance prediction.

\subsection{Experimental Setup}

\textbf{Dataset}: We use the open-source driving simulator CARLA\cite{dosovitskiy2017carla} for generating our driving dataset. 
We collect the training data by driving the agent vehicle using the CAL controller \cite{sauer2018conditional} with the ground truth affordances as input.
The collected data comprises of 262 driving episodes and with a total of 100,000 video frames.
For each video frame, we collect: (1) front center camera image, and (2) six affordances (Discrete: Red Light, Hazard Stop, Speed Sign ; Continuous: Relative Angle, Centerline Distance, Vehicle Distance) \cite{sauer2018conditional}. We used approximately $85\%$ of the video frames as the training data, and the remaining as test set.



\begin{figure}[h!]
    \centering
    \includegraphics[width=0.8\columnwidth,height=4cm]{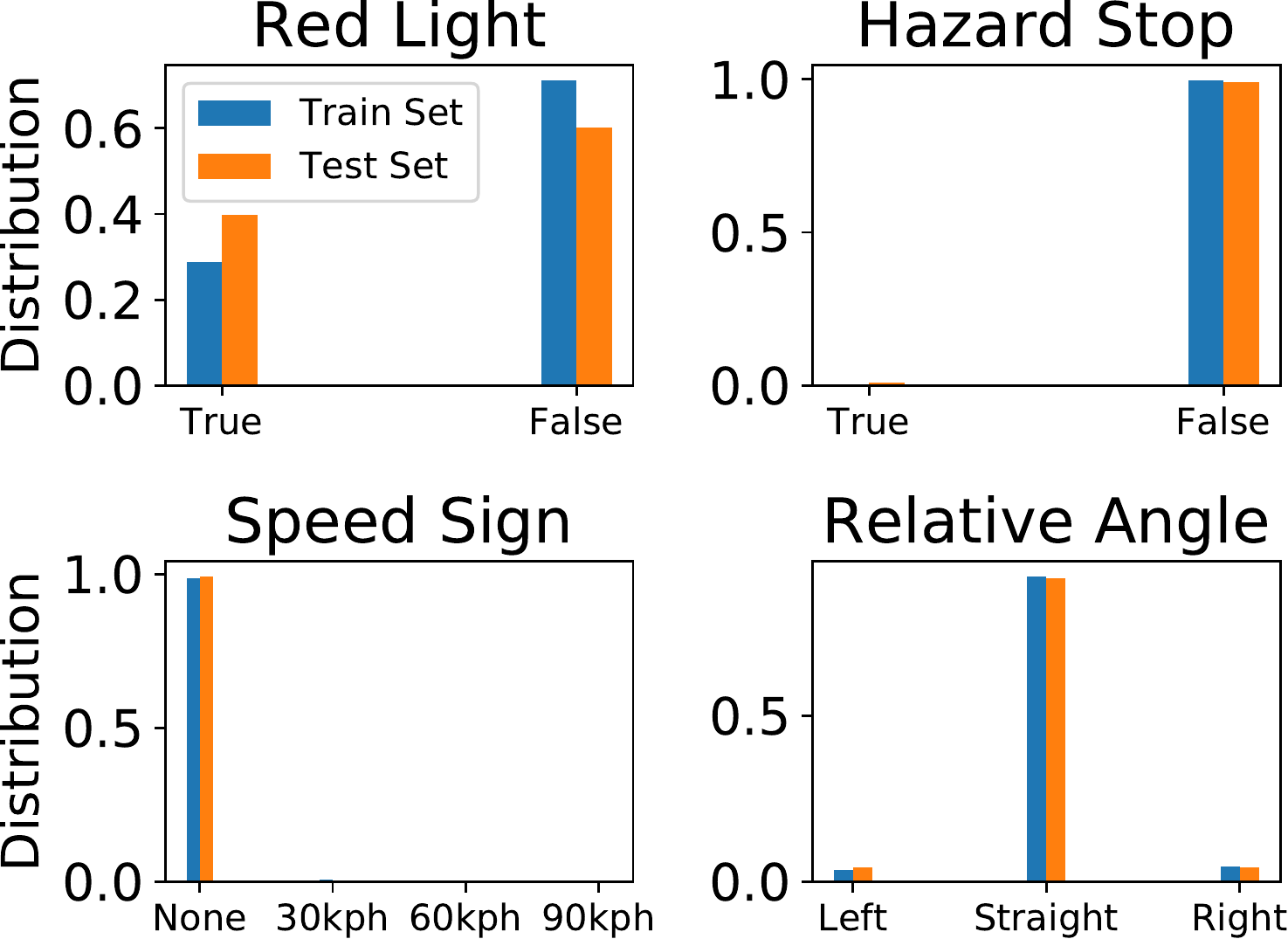}
    \label{fig:red}
    \caption{\textbf{Distribution of affordances in train and test set}} 
    \label{fig:distrib}
\end{figure}

 Figure \ref{fig:distrib} illustrates the distribution of four affordances per class label, in training and test set. Red light and Hazard Stop have two classes - \textit{True}, \textit{False}, Speed sign has four classes - \textit{None} , \textit{30kmph}, \textit{60kmph}, \textit{90kmph}, while in case of Relative angle which is the orientation angle of the agent vehicle, we have grouped the values into buckets corresponding to \textit{Left turn}, \textit{Straight} and \textit{Right turn}. We can divide the entire range of Relative Angle affordance across 20 buckets ranging from -1.0 to +1.0. Note that buckets corresponding to (-1.0 to -0.1), (-0.1 to 0.1) and (0.1 to 1.0) approximately indicate the left turn, straight road and right turn respectively.
 
\textbf{Subset selection on Autonomous Driving:}
We use Conditional Affordance Learning \cite{sauer2018conditional} (CAL) model as the driving model. We carry out training on NVIDIA Tesla P100 for 50 epochs using Adam Optimizer with learning rate varying between $10^{-3}$ to $10^{-4}$. We use SIFT dissimilarity as the pairwise metric ($d_{ij}$) for all selection methods and two variations of losses ($L_j$) as pointwise metric - total loss (TL), bucket specific relative angle loss aided with other multi-task losses (BML). We observe that among all 6 affordances, relative angle which provides the steering angle of the car, is the most essential affordance for episode completion.So,for pointwise metric, we provide the highest weightage to the losses corresponding to buckets of relative angle meant for turns and intersection, followed by the other affordances. We report results with the following methods of selection for compression ratios 100:20 and 100:7:

\begin{enumerate}
    \item \textbf{Whole Set (WS)}: Unreduced dataset.
    \item \textbf{Uniform Skip (US)}: Uniformly sampled data.
    \item \textbf{Only SIFT (OS)}: Only $d_{ij}$.
    \item \textbf{Only Loss (OL)}: Images with highest losses.
    \item \textbf{\prl}\cite{das2020multi} : $d_{ij}$ with TL.
    \item \textbf{{\oursubmodular-BML}(SBML)} :  $d_{ij}$ with BML.
    \item \textbf{\newformulation-TL(TCTL)} : $d_{ij}$ with TL.
    \item \textbf{\newformulation-BML(TCBML)} : $d_{ij}$ with BML.
\end{enumerate}








 \textbf{Subset selection and Training Times}:
Table \ref{tab:time} shows the amount of total time taken for training and subset selection for each compression ratio. Compression ratio of 100:20 takes about 1/3rd the total time taken to train the whole set (WS). 
This follows for the other compression ratio too and hence saves both time and space, essential in the current IoT setting with massively huge data.

\begin{table}[]
\caption{\textbf{Time complexity of subset selection methods}}
\footnotesize
\begin{tabular}{|l|l|l|l|l|}
\hline
\begin{tabular}[c]{@{}l@{}}Compression\\ Ratio\end{tabular}                         & WS                 & \multirow{2}{*}{}                                         & 100:20 & 100:7 \\ \cline{1-2} \cline{4-5} 
\begin{tabular}[c]{@{}l@{}}Training Time\\ (hours)\end{tabular}                     & 42                 &                                                           & 11     & 7     \\ \hline
\multirow{3}{*}{\begin{tabular}[c]{@{}l@{}}Subset Selection\\ (hours)\end{tabular}} & \multirow{3}{*}{0} & MCOSS                                                       & 2      & 1     \\ \cline{3-5} 
                                                                                    &                    & SBML                                                      & 5      & 2.5   \\ \cline{3-5} 
                                                                                    &                    & \begin{tabular}[c]{@{}l@{}}Proposed\\ Method\end{tabular} & 1.5    & 0.75  \\ \hline
\end{tabular}
\label{tab:time}
\end{table}

\textbf{Metrics reported}: We have reported metrics for the two subtasks: episode completion and affordance accuracies. An episode corresponds to a source-destination pair provided to the agent vehicle. The episode completion is measured in terms of number of successfully completed episodes. The accuracies of discrete affordances are measured in terms of Micro Accuracy (MIC) and Macro Accuracy (MAC), while the continuous affordances are measured in terms of Mean Absolute Error (MAE). 

\textbf{Affordance Selection}: We have reported results for five affordances - \textit{Red light}, \textit{Hazard Stop}, \textit{Relative Angle}, \textit{Centerline Distance} and \textit{Vehicle Distance}. Considering the setup of our dataset collection, from Figure \ref{fig:distrib}, we can observe that the two affordances \textit{Speed Sign} and \textit{Hazard Stop} suffer from class imbalance. Alongside, prediction accuracy of \textit{Speed Sign} does not have an impact on episode completion. Hence, we refrain from reporting its accuracy. 



\subsection{Comparison of approximate online subset selection methods}

\vspace{-2.7mm}
\begin{figure}[h!]
    \centering
    \begin{subfigure}{0.49\columnwidth}
    \centering
    \includegraphics[width=\textwidth,height=2.5cm]{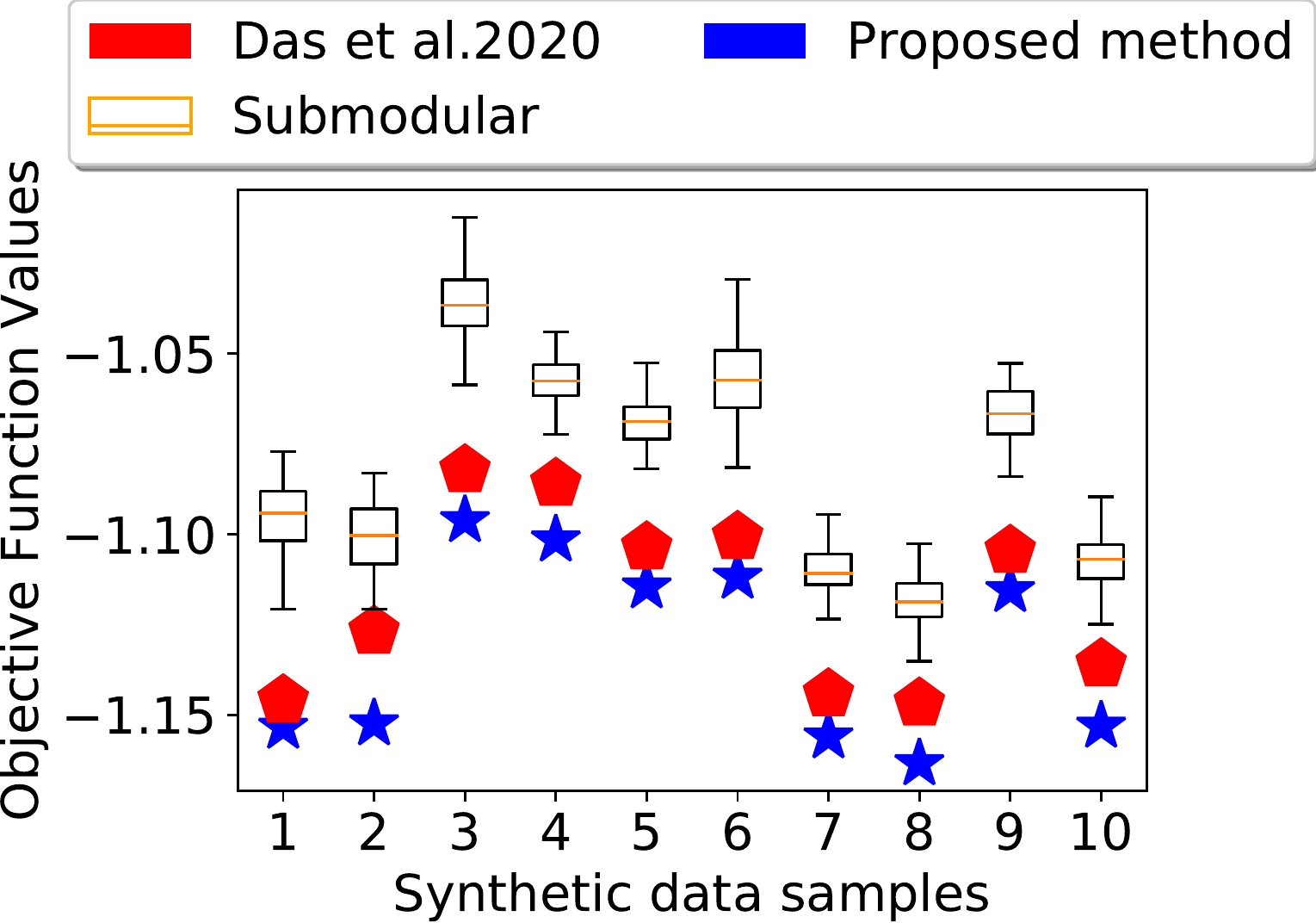}
    \label{fig:synbox}
    \end{subfigure}
    \begin{subfigure}{0.49\columnwidth}
     \centering
     \includegraphics[width=\textwidth,height=2.5cm]{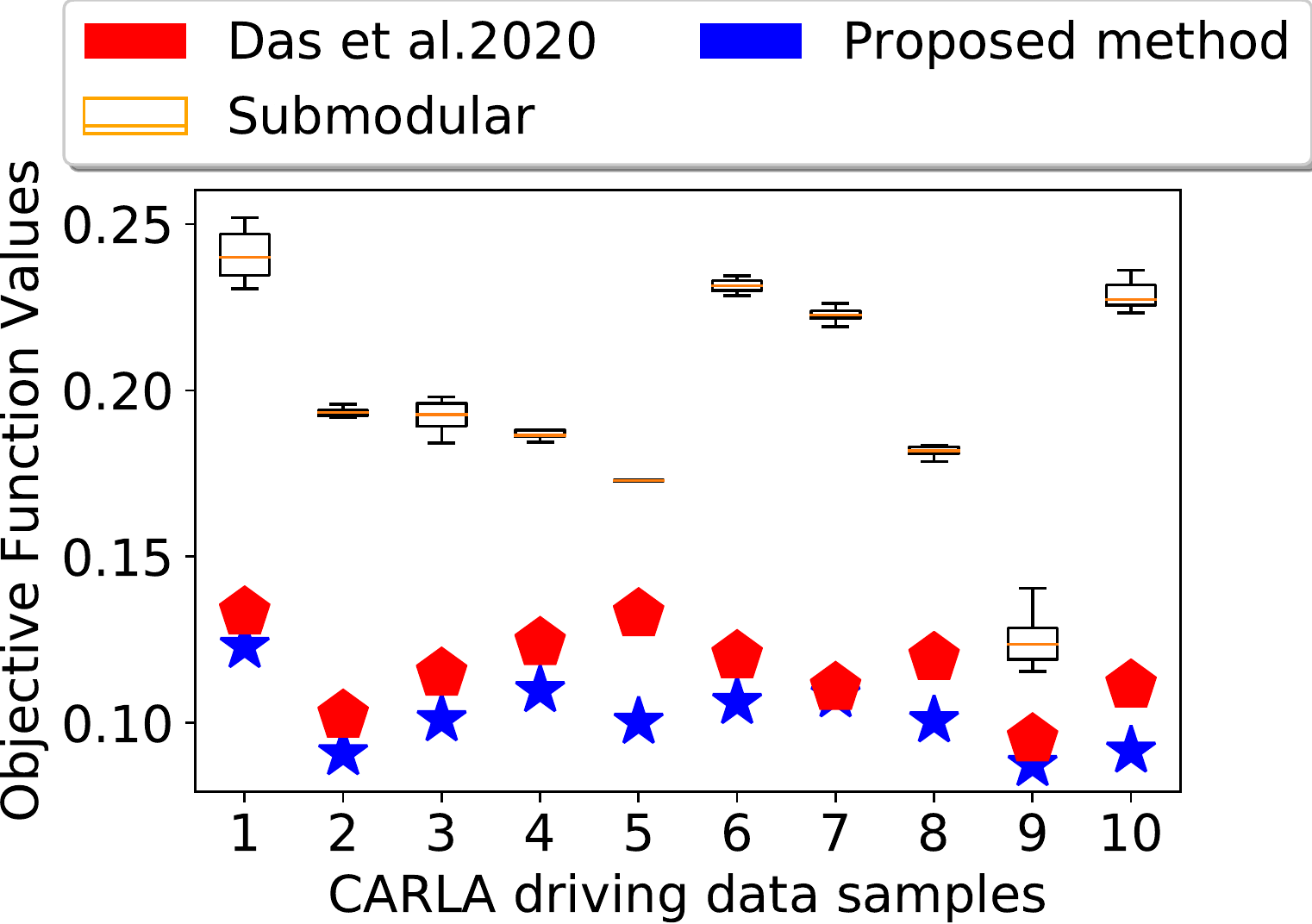}
    \label{fig:realbox}
     \end{subfigure}
     \caption{\textbf{Objective function values for proposed method , \cite{das2020multi} method and Submodular method for (left) Synthetic data and (right) CARLA driving data samples}}
     \label{fig:convsub}
 \end{figure}  

In this section, we compare the optimal subsets reported by the baseline method \cite{das2020multi}, submodular method and the proposed convex optimization based method in terms of final objective function values. 
Figure \ref{fig:convsub}a shows the objective function values for the three approaches, for 10 randomly synthesized problem instances ($d$ matrix of dimension $100\times 100$ and $L$ vector of dimension $100$). For the submodular method, we report a box plot of results over 100 runs of the algorithm to capture the randomness. While all three methods find approximate solutions, the proposed convex method consistently finds lower values of objective function, followed by \cite{das2020multi}, and submodular optimization. For real data (CARLA data for our further experiments), we report the same in Figure \ref{fig:convsub}b , for 10 episodes in our collected data. We can clearly observe that function values attained by our proposed method lies below that of the other approximate methods, thus proving our method to be an efficient approach.

Next, we discuss the application of the subset selection methods on autonomous driving.

\begin{table*}[]
\scriptsize
\centering
\captionsetup{justification=centering}
\caption{\textbf{Comparison of data gradation techniques in terms of episode completion in training and test conditions for 100:20 and 100:7 compression ratio}}
\label{tab:ep5_15}

\begin{tabular}{|l|l|l|l|l|l|l|l|l|l|}
\hline
\multirow{2}{*}{\begin{tabular}[c]{@{}l@{}}Compression\\ Ratio\end{tabular}} & \multirow{2}{*}{Methods}                               & \multicolumn{4}{c|}{\begin{tabular}[c]{@{}c@{}}Training\\ Conditions\end{tabular}}                                                            & \multicolumn{4}{c|}{\begin{tabular}[c]{@{}c@{}}Test\\ Conditions\end{tabular}}                                                                \\ \cline{3-10} 
                                                                             &                                                        & Straight & One-Turn & \begin{tabular}[c]{@{}l@{}}Straight\\ Dynamic\end{tabular} & \begin{tabular}[c]{@{}l@{}}One-Turn\\ Dynamic\end{tabular} & Straight & One-Turn & \begin{tabular}[c]{@{}l@{}}Straight\\ Dynamic\end{tabular} & \begin{tabular}[c]{@{}l@{}}One-Turn\\ Dynamic\end{tabular} \\ \hline
                                                                             & WS    & 10       & 10                                                         & 10        & 10                                                          & 10       & 10                                                         & 10        & 10                                                          \\ \hline
\multirow{4}{*}{100:20}                                                      & US & 9        & 3                                                          & 8        & 3                                                          & 9        & 5                                                          & 9        & 5                                                          \\ \cline{2-10} 
                                                                             & OS                                                   & 10       & 7                                                          & 10       & 7                                                          & 10       & 7                                                          & 9        & 6                                                          \\ \cline{2-10} 
                                                                            & OL                                                   & 10       & 6                                                          & 10       & 6                                                          & 10       & 7                                                          & 9        & 7                                                          \\ \cline{2-10} 
                                                                             & MCOSS                                                    & 9        &  8                                                         & 9        & 8                                                          & 7       &  7                                                         & 7        & 7                                                          \\ \cline{2-10} 
                                                    & SBML                                                    & 9        &  7                                                         & 9        & 7                                                          & 9       &  7                                                         & 9        & 7                                                          \\ \cline{2-10} 
                                                                             & TCTL                                                    & 10       & 8                                                         & 10       & 7                                                          & 10       & 9                                                          & 10       & 9 
                                \\ \cline{2-10} 
                                                                             & TCBML                                                    & 10       & \textbf{10}                                                          & 10       & \textbf{10}                                                          & 10       & \textbf{10}                                                          & 10       &\textbf{10}                                \\        \hline
                                                        \hline
\multirow{4}{*}{100:7}                                                      & \begin{tabular}[c]{@{}l@{}}MCOSS\end{tabular} & 9        & 5                                                            & 9        & 5                                                          & 7        &  4                                                         & 7        & 4                                                          \\ \cline{2-10}
                                                                             & SBML                                                    & 9        & 3                                                            & 9        & 3                                                          & 9        & 2                                                          & 9        & 2                                                          \\ \cline{2-10}
                                                                             & TCTL                                                    & 10       & 7                                                          & 10        & 7                                                          & 10       & \textbf{9}                                                          & 10       & \textbf{9}  
                                \\ \cline{2-10} 
                                                                             & TCBML                                                    & 10       & \textbf{8}                                                           & 10       & \textbf{8}                                                          & 10       &\textbf{9}                                                           & 10       & \textbf{9}                                 \\        \hline                                                    
\end{tabular}
\end{table*}

\subsection{Comparison of Methods: Episode Completion}

In this section, we will study the performance of different subset selection techniques on the basis of episode completion which is the main intended objective.

We consider four tasks under episode completion which had been originally defined in \cite{dosovitskiy2017carla}:
\begin{enumerate}
    \item Straight : The episode comprises of pairs of points which are straight roads.
    \item One-Turn : The episode comprises of pairs of points which include left or right turn in their paths.
    \item Straight Dynamic : Similar to \textit{Straight} task, it occurs in the presence of other vehicles and pedestrians.
    \item One-Turn Dynamic : Similar to \textit{One-Turn} task, it occurs in the presence of other vehicles and pedestrians.
\end{enumerate}

We show in Table \ref{tab:ep5_15} the performance of CAL model\cite{sauer2018conditional}  trained on subsets, obtained by various selection techniques, by simulating it with the CARLA \cite{dosovitskiy2017carla} simulator.

We report number of successfully completed episodes (out of a total of 10 episodes) for each subtask in training and test conditions. We observe that the tasks - \textit{Straight} and \textit{Straight Dynamic} are fairly easy to accomplish for all subset selection methods. The challenge arises in the other two tasks where we see a large number of unsuccessful episodes.

The completion of episodes in turns, largely depends on the affordance \textit{Relative Angle} which as we mentioned, provides the steering angle of the agent vehicle. We observe that Uniform Skip (US) is unable to complete majority of the turns in training or test conditions. This is due to the inherent nature of selection of US method, which does not take the important frame sequences into consideration.

\begin{figure*}[h!]
     \centering
    \begin{subfigure}{0.65\columnwidth}
    \centering
    \includegraphics[width=\textwidth,height=3cm]{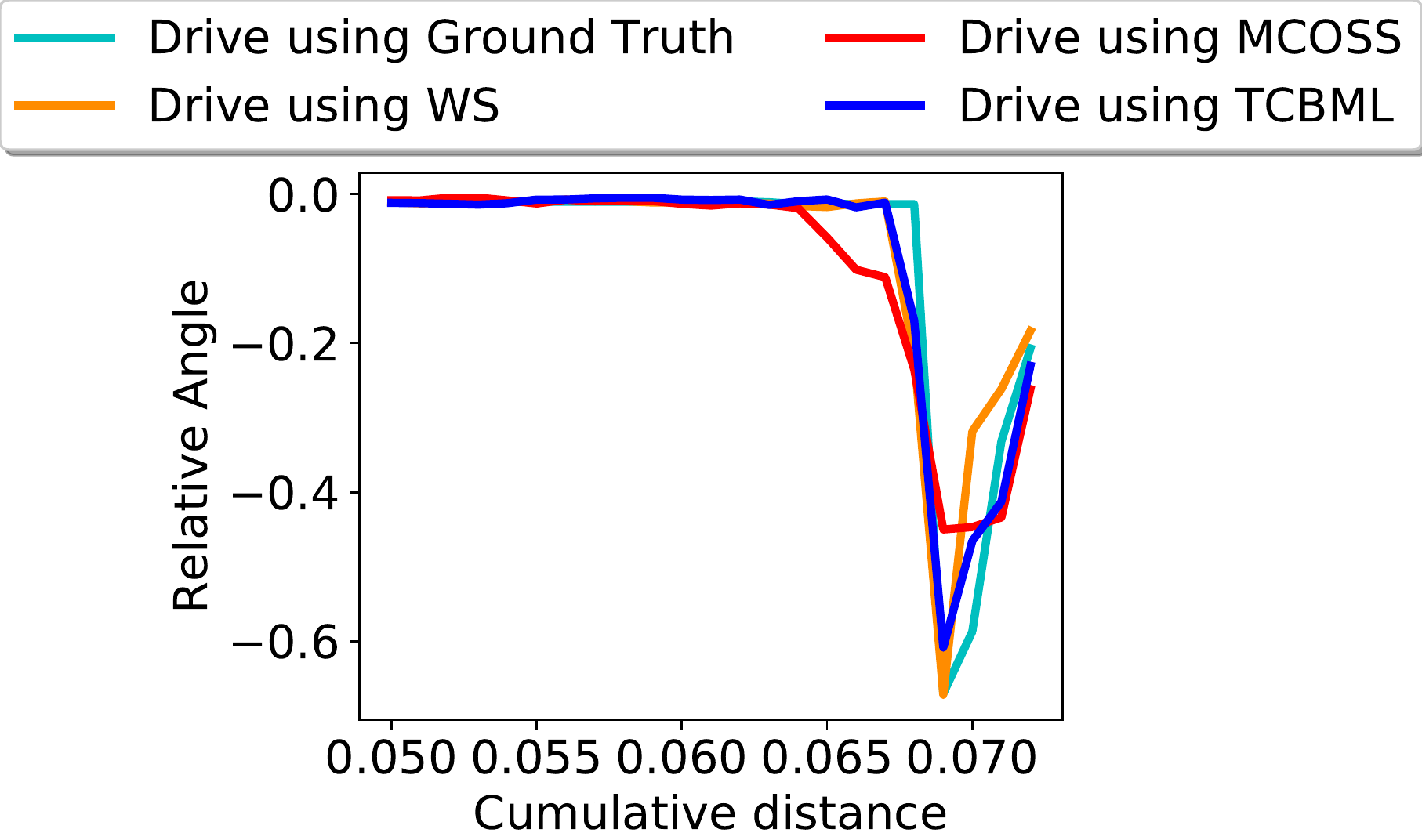}
    \label{fig:tskip15}
    \end{subfigure}
    \centering
    \begin{subfigure}{0.65\columnwidth}
    \centering
    \includegraphics[width=\textwidth,height=3cm]{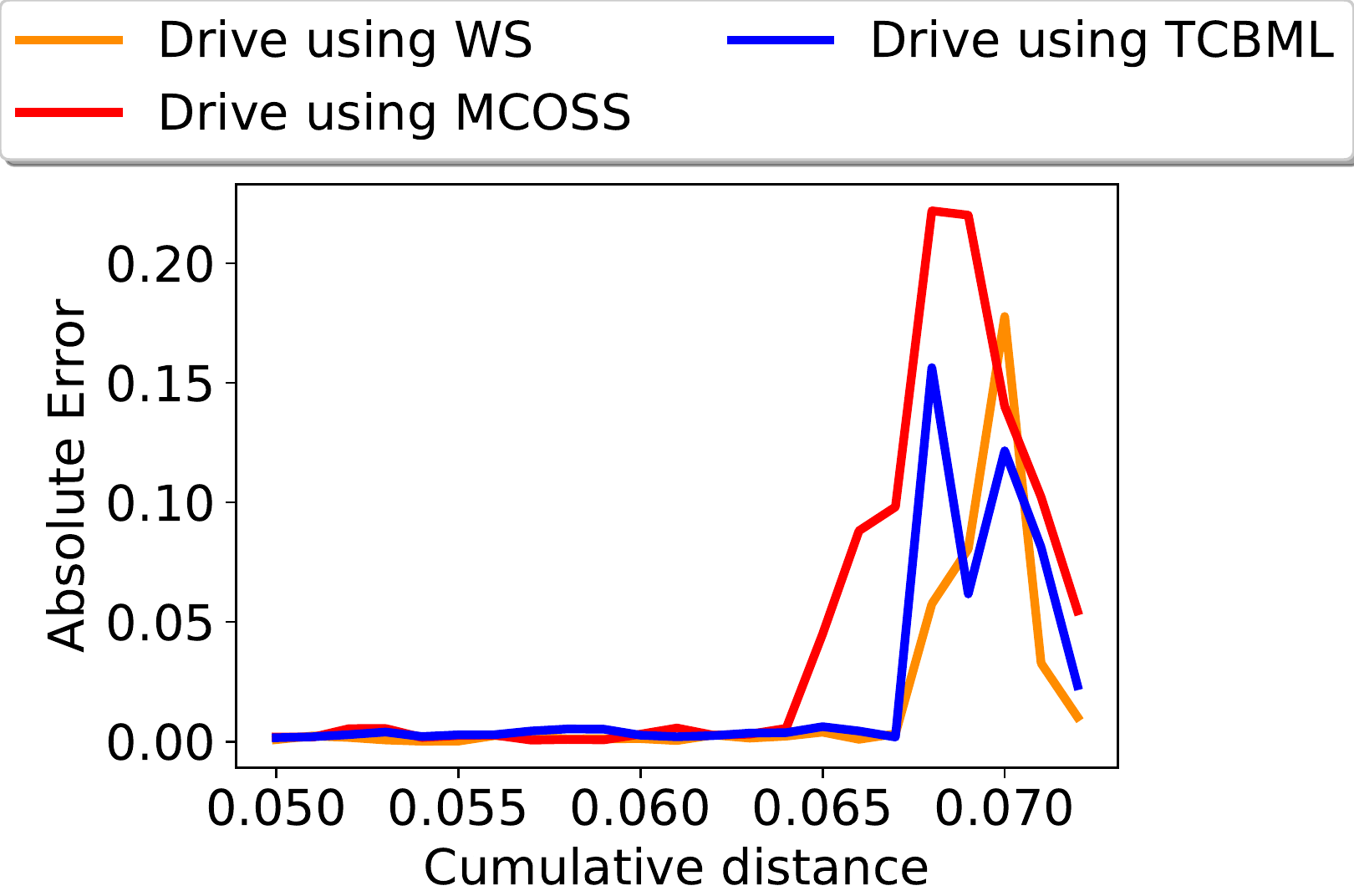}
    \label{fig:tskip5}
    \end{subfigure}
    \centering
    \begin{subfigure}{0.65\columnwidth}
    \centering
    \includegraphics[width=\textwidth,height=3cm]{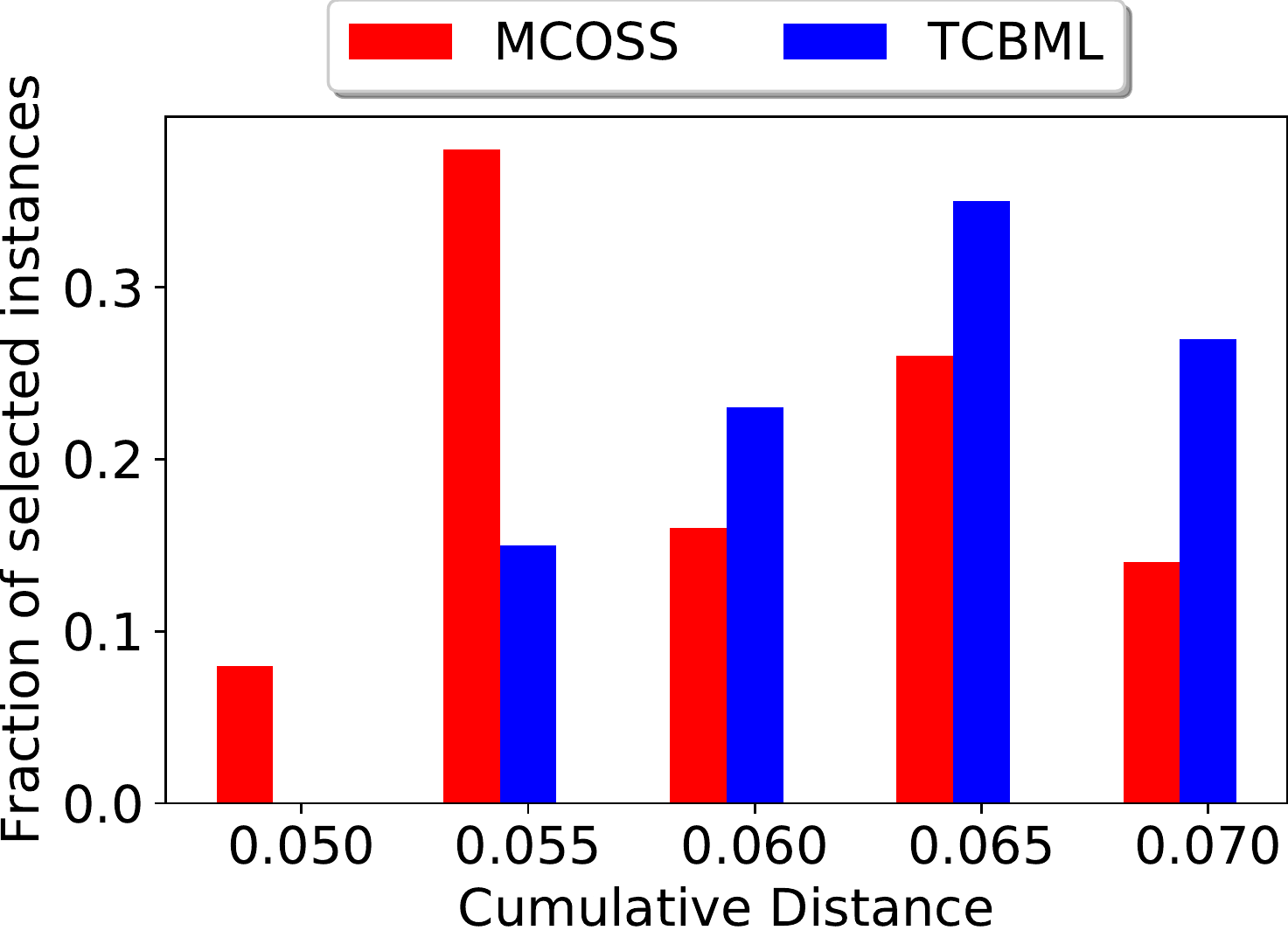}
    \label{fig:tskip5}
    \end{subfigure}
    \caption{\textbf{Anecdotal example of a sample episode showing (1)the region of failure by MCL (2) the corresponding errors and (3) fraction of selected instances from those regions. CBML selects more fraction of instances from regions important for episode completion.}}
    \label{fig:deviate}
\end{figure*}

On the other hand, OS, OL, MCOSS, SBML complete some of the turn episodes, with TCBML completing the highest of all methods (100\% both in training and test conditions) followed by TCTL.
This is due to the efficient usage of combination of pointwise metric along with pairwise metric in the selection of instances.
We also report the episode completion numbers for 100:7 compression ratio in Table \ref{tab:ep5_15}, for MCOSS and SBML and the thresholded convex methods. We can observe that TCTL and TCBML continue to perform better compared to the other baselines with TCBML completing 80\% and 90\% episodes in training and test conditions even after dropping 93\% of the frames.

We show an anecdotal example of a failed episode from MCOSS \cite{das2020multi} method in Figure \ref{fig:deviate} and try to analyse the reason behind it. In Figure \ref{fig:deviate}, we firstly show a plot of an episode with distance covered vs Relative Angle affordance. The cumulative distance is chosen as an axis of reference in order to make the comparison spatially similar for all the methods. Overall, the agent vehicle in the simulator is resilient to errors. However, we can observe the gradual deviation of MCOSS \cite{das2020multi} method and its accumulated errors which lead to an incomplete episode. We can also relate the region of failures to the corresponding relative angle errors in Figure \ref{fig:deviate} (\textit{middle}) where we observe that overall, MCOSS \cite{das2020multi} is having a higher error compared to proposed method TCBML and model trained on WS. 
We also show the fraction of selected instances by both the selection methods, in Figure \ref{fig:deviate} (\textit{right)} from the same region, and we can observe that TCBML selects more fraction of instances in the area right before the turn and during the turn, which eventually help in its completion.
We discuss the next subtask - affordance prediction in the following subsection.

\begin{table}[]
\footnotesize
\caption{\textbf{Comparison of data gradation techniques for 100:20 compression ratio on Hazard Stop, Red Light, Vehicle Distance and Centerline Distance}}
\label{tab:allaff}
\begin{tabular}{|l|p{8 mm}|p{8 mm}|p{8 mm}|p{8 mm}|l|l|}
\hline
\multicolumn{1}{|c|}{\multirow{2}{*}{Method}} & \multicolumn{2}{c|}{Hazard Stop}                                                                                    & \multicolumn{2}{c|}{Red Light}                                                                                       & \multicolumn{1}{c|}{\begin{tabular}[c]{@{}c@{}}Vehicle\\ Dist\end{tabular}} & \multicolumn{1}{c|}{\begin{tabular}[c]{@{}c@{}}Center\\ Dist\end{tabular}} \\ \cline{2-7} 
\multicolumn{1}{|c|}{}                        & \multicolumn{1}{c|}{MIC}                                 & \multicolumn{1}{c|}{MAC}                                 & \multicolumn{1}{c|}{MIC}                                 & \multicolumn{1}{c|}{MAC}                                  & \multicolumn{1}{c|}{MAE}                                                        & \multicolumn{1}{c|}{MAE}                                                           \\ \hline
WS                                            & 99.44                                                    & 99.72                                                    & 97.39                                                    & 97.41                                                     & 0.03                                                                            & 0.05                                                                               \\ \hline
US                                            & 99.14                                                    & 99.47                                                    & 96.72                                                    & 97.03                                                     & 0.06                                                                            & 0.05                                                                               \\ \hline
OS                                            & 99.11                                                    & 99.56                                                    & 96.11                                                    & 96.07                                                     & 0.09                                                                            & 0.08                                                                               \\ \hline
OL                                            & 99.41                                                    & 99.6                                                     & 92.38                                                    & 91.08                                                     & 0.09                                                                            & 0.08                                                                               \\ \hline
MCOSS                                         & 99.43                                                    & 98.39                                                    & 91.63                                                    & 89.73                                                     & 0.05                                                                            & 0.06                                                                               \\ \hline
SBML                                          & 99.25                                                    & 99.60                                                    & 94.33                                                    & 93.91                                                     & 0.05                                                                            & 0.06                                                                               \\ \hline
TCTL                                          & 99.31                                                    & 99.61                                                    & 96.38                                                    & 96.05                                                     & 0.05                                                                            & 0.05                                                                               \\ \hline
TCBML                                         & \begin{tabular}[c]{@{}l@{}}99.45\\ (0.01\%)\end{tabular} & \begin{tabular}[c]{@{}l@{}}99.71\\ (0.01\%)\end{tabular} & \begin{tabular}[c]{@{}l@{}}93.98\\ (3.41\%)\end{tabular} & \begin{tabular}[c]{@{}l@{}}92.83)\\ (4.58\%)\end{tabular} & \begin{tabular}[c]{@{}l@{}}0.04\\ (33.3\%)\end{tabular}                        & \begin{tabular}[c]{@{}l@{}}0.05\\ (0\%)\end{tabular}                               \\ \hline
\end{tabular}
\end{table}

\subsection{Comparison of Methods: Affordance Accuracies}

\begin{figure*}[h!]
     \centering
    \begin{subfigure}{0.5\columnwidth}
    \centering
    \includegraphics[width=\textwidth,height=3cm]{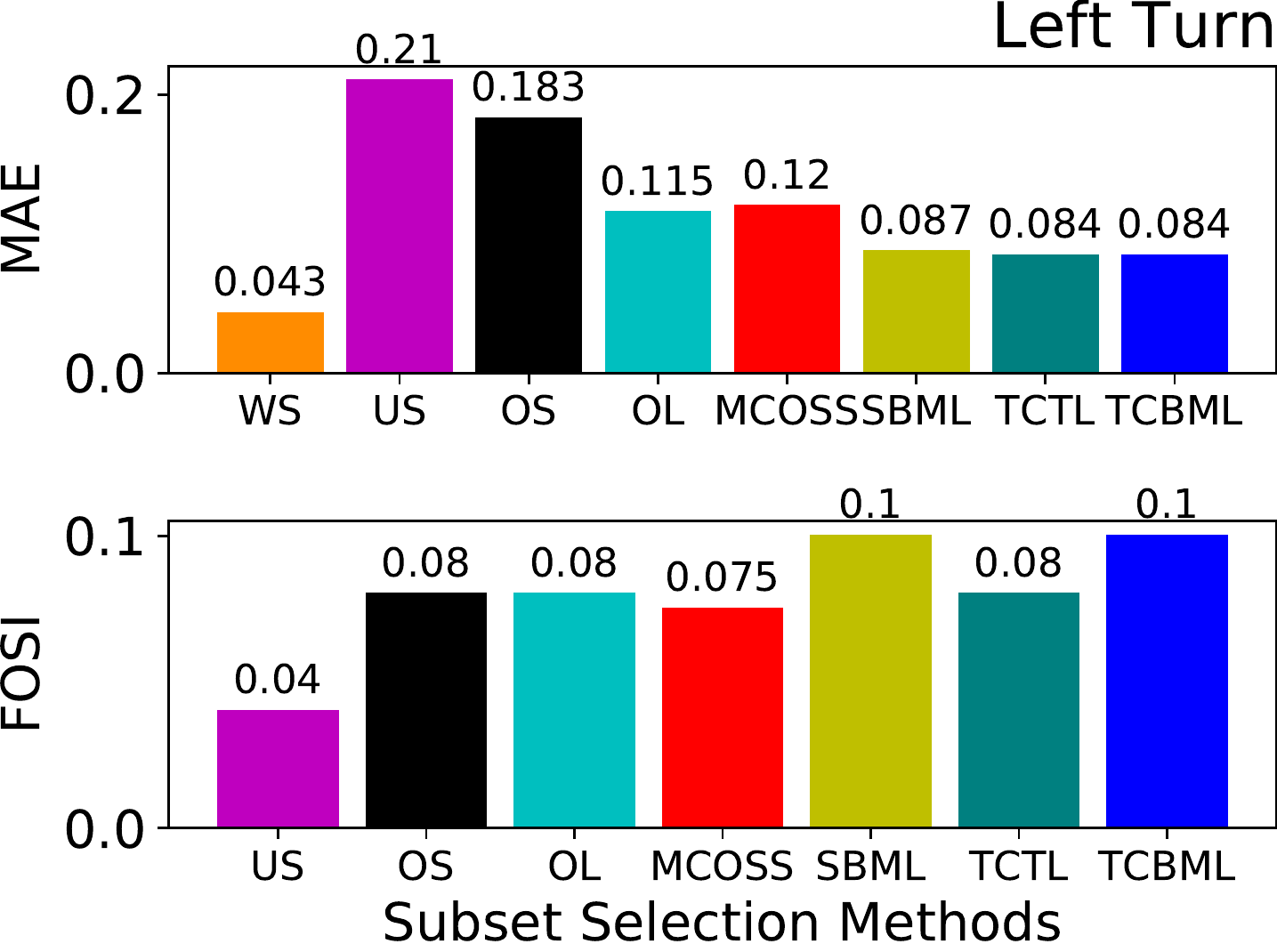}
    \label{fig:tskip15}
    \end{subfigure}
    \centering
    \begin{subfigure}{0.5\columnwidth}
    \centering
    \includegraphics[width=\textwidth,height=3cm]{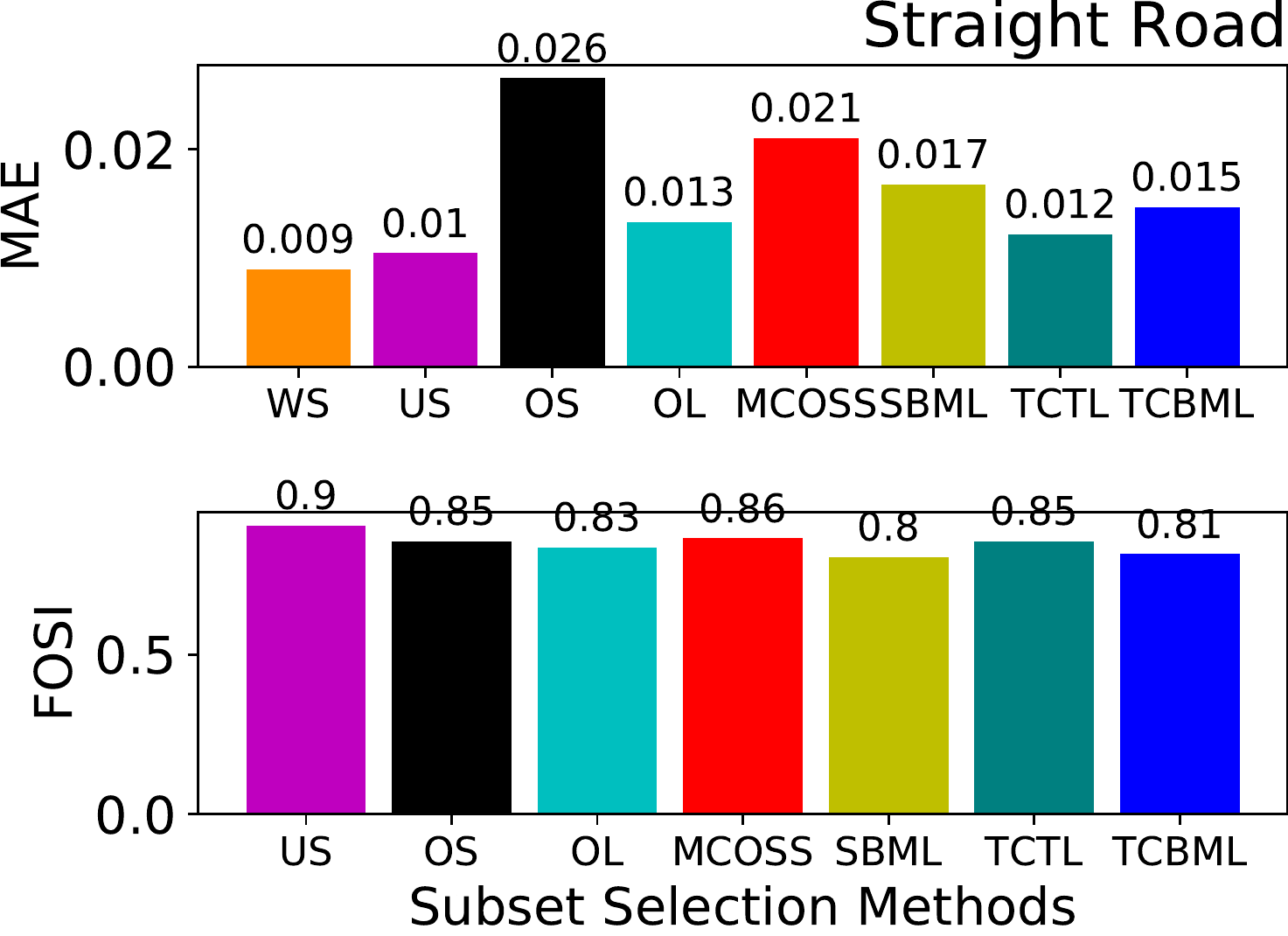}
    \label{fig:tskip5}
    \end{subfigure}
    \centering
    \begin{subfigure}{0.5\columnwidth}
    \centering
    \includegraphics[width=\textwidth,height=3cm]{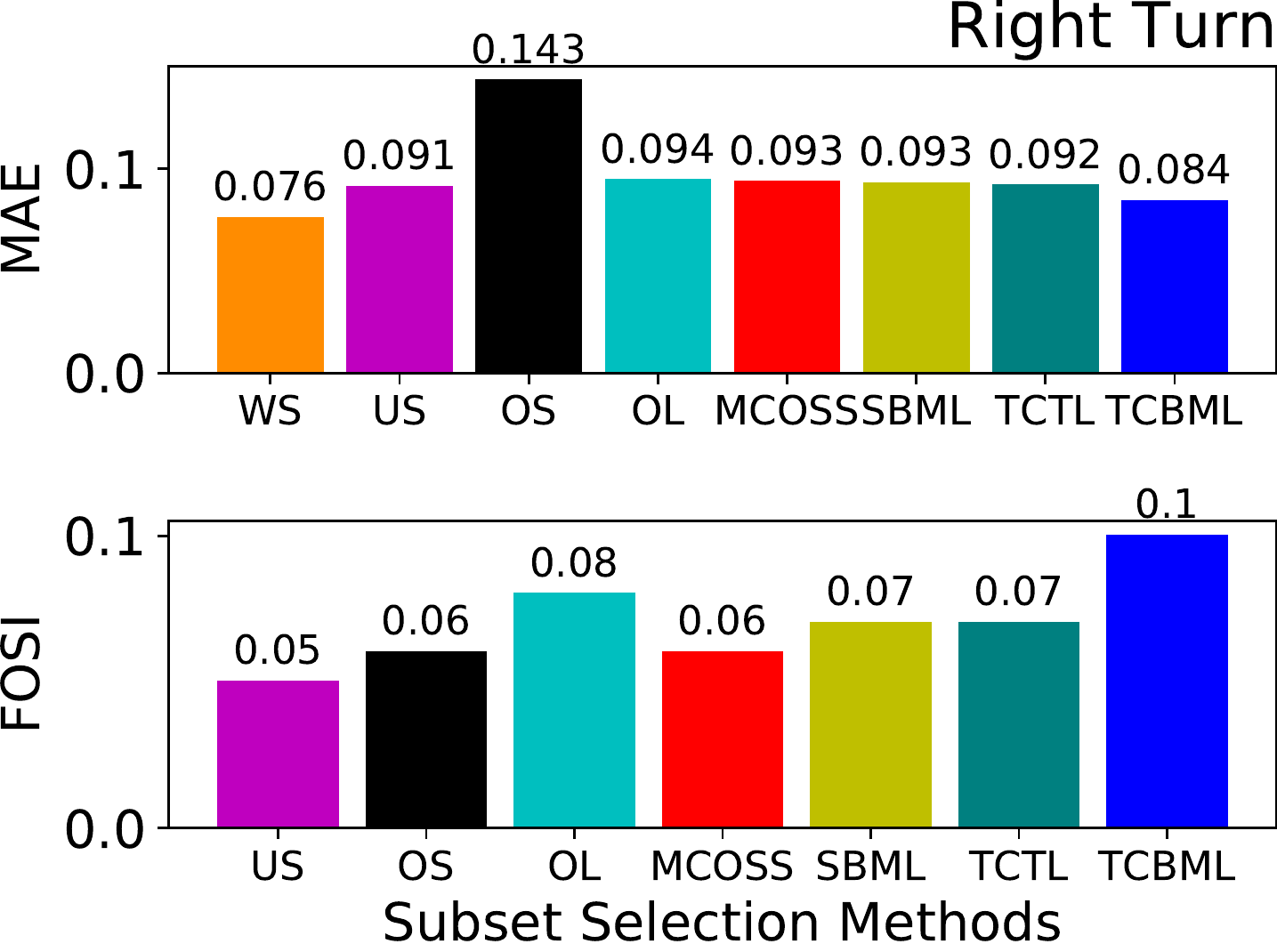}
    \label{fig:tskip5}
    \end{subfigure}
    \caption{\textbf{Fraction of selected instances (FOSI) and Mean Absolute Error (MAE) for Relative Angle for all selection methods. More fraction of instances in turn buckets lead to lower MAE in CBML for 100:20 compression ratio.}}
    \label{fig:fracmae_100_20}
\end{figure*}

In this section, we will study the performance of different subset selection methods on the basis of their performance in affordance predictions.

We show in Table \ref{tab:allaff} the performance of CAL \cite{sauer2018conditional} driving model for four affordances, on subsets obtained by different subset selection techniques for 100:20 compression ratio. We also show the \% difference in performance of our proposed method TCBML with that of WS.
We notice that all the methods can classify the affordances reasonably well for both red light and hazard stop affordance. This is due to weighted cross-entropy losses used for the large number of imbalanced classes as seen initially in Figure \ref{fig:distrib}. We can also see that TCTL and TCBML outperform all other methods in predicting vehicle distance affordance. In case of centerline distance, all the methods work similarly with OS and OL performing the worst.

In case of \textit{Relative Angle} affordance, we have grouped the values into three different buckets meant for Left Turn, Straight Road, Right Turn. We observe in Figure \ref{fig:fracmae_100_20} that the MAE for straight road buckets for all selection methods are very low in range. This is essentially due to the skewness in Relative Angle affordance. However, TCTL and TCBML outperform all other methods in turn buckets which have in usual, lesser number of datapoints. We can also observe that TCBML selects more fraction of instances for both Left and Right turns while relaxes comparatively in the Straight Road. SBML also has close fraction of instances to TCBML for left turn, however it drops down in the fraction of right turns. 
The difference in MAE becomes more evident with increase in compression ratio. We have added the result for 100:7 in the Supplementary.

This is indicative of the fact that the proposed  thresholded convex method gives more importance to the frames which are non-redundant and also having higher losses, thus leading to better performance. Thus, we can clearly see that our proposed methods TCTL and TCBML are performing better quantitatively in terms of affordance predictions and qualitatively in terms of deployment of the model.

\section{Conclusion}
In this paper, we propose a novel thresholded convex optimisation based online video frame subset selection technique involving pairwise and pointwise similarity for optimal learning for the task of self-driving. We also propose a natural set-function based criteria to incorporate pointwise criteria using submodular optimisation. We show the effectiveness of our improved subset selection method (TCBML and TCTL) in the aspect of episode completion and affordance predictions. We did extensive experiments to show that even after dropping 80\% of frames, we succeed in completing all the episodes involving the difficult task of turns.


{\small
\bibliographystyle{ieee_fullname}
\bibliography{egbib}
}

\end{document}


\title{Convex Online Video Frame Subset Selection using Multiple Criteria for Data Efficient Autonomous Driving : Supplementary Material}

\author{First Author\\
Institution1\\
Institution1 address\\
{\tt\small firstauthor@i1.org}
\and
Second Author\\
Institution2\\
First line of institution2 address\\
{\tt\small secondauthor@i2.org}
}

\maketitle

\section{Proof of submodularity}
For every set $S$, $f(S)$ can be defined as:
\small
\begin{equation}
    f(S) = \sum_{i \in X} min \{ min_{j \in R} Q_{ij} , min_{j \in S} Q_{ij} \} 
\label{eq:submod}
\end{equation}
\normalsize

\begin{remark}
-f(S) is submodular.
\end{remark}

\textbf{Proof:} According to definition of submodularity, for $-f(S)$ to be submodular, 
\small
\begin{equation}
    -f(S \cup \{x\}) - (-f(S)) \geq -f(T \cup \{x\}) - (-f(T))
\end{equation}
\normalsize
where $S \subseteq T$ , $x \notin S $ , $x \notin T $. Let us define the notations:\\
\small
\begin{enumerate}
    \item $D_i(S)$ : Function value for finding a representative for an element i $\in$ X with past representative set as $S$. 
    \item $D_i(T)$ : Function value for finding a representative for an element i $\in$ X with past representative set as $T$.
    \item $D_{ix}$ : Function value for finding a representative for an element i $\in$ X with past representative set as $S \cup x$ or $T \cup x$.
\end{enumerate}
\normalsize
Note that there can be 3 possibilities after addition of $x$.
\small
\begin{enumerate}
    \item $x$ does not become a representative of $i \in X$.
    \item $x$ becomes a representative of $i \in X$ only with past representative set as $S$.
    \item $x$ becomes a representative of $i \in X$ with both $S$ and $T$ as past representative sets. 
\end{enumerate}
\normalsize
To prove: 
\small
\begin{equation}
    f(S) - f(S \cup \{x\}) \geq f(T) - f(T \cup \{x\})
\end{equation}
\normalsize
or
\small
\begin{equation}
    D_i(S) - D_{ix} \geq  D_i(T) - D_{ix}
\end{equation}
\normalsize
Under the first possibility, both sides of inequality lead to $0$.

Since $S \subseteq T$, by the form of $f(S)$, $D_i(T) \leq D_i(S)$. Addition of representative for an element $i \in X$ can either decrease the function value or keep it the same.
\small
\begin{equation}
    \therefore{} D_i(T) - D_{ix} \geq 0 , 
    D_i(S) - D_{ix} \geq 0
\end{equation}
\normalsize
Hence, under the second possibility, $D_i(S) - D_{ix} \geq  D_i(T) - D_{ix}$ or $D_i(S) - D_{ix} \geq  0$.

Now, $\because D_i(T) \leq D_i(S), \therefore{} D_i(T) - D_{ix} \leq D_i(S) - D_{ix}$ or $D_i(S) - D_{ix} \geq D_i(T) - D_{ix}$.

Therefore, $-f(S)$ is submodular.

\section{Conditions for MCOSS}
The objective function for MCOSS is:
\begin{align}
\label{eqn:oldsiftloss}
    &\min_{z^o_{ij},z^n_{ij}} \sum_{i=1}^m \sum_{j=1}^{|R_t|} z_{ij}^{o} Q^o_{ij} + \sum_{i,j=1}^m z_{ij}^n Q^n_{ij} + \lambda \sum_{j=1}^m \| [z_{1,j}^n \dots z_{m,j}^n ] \|_p \nonumber \\ 
    &s.t.\sum_{j=1}^{|R_t|} z_{i,j}^o + \sum_{j=1}^m z_{i,j}^n = 1, \ \forall i\in X_{t+1} \nonumber \\
    & z_{i,j}^n, z_{i,j}^o \in [0,1], \ \forall i,j
\end{align}

\begin{theorem}
\label{thm:representative-condition}
Let $z_{ij}^{o}$ and $z_{ij}^{n}$ be the optimal solution for formulation \ref{eqn:oldsiftloss}.
A new frame $j\in X_{t+1}$ is selected as a representative frame for at least one incoming frame $i\in X_{t+1}$, i.e. $z_{ij}^n = 1$, only if the following conditions hold:
\begin{itemize}
\item For some incoming frame  $i \in X_{t+1}$ , $Q^n_{ij} < Q^n_{ij'}$, for all $j'\in X_{t+1}$ and $j' \neq j$
\item For some incoming frame  $i \in X_{t+1}$, $ Q^n_{ij} < \frac{ \sum_{i'=1}^m z_{i',k}^o Q^o_{i'k} + \lambda \| [z_{1,j}^n \dots z_{m,j}^n ] \|_p }{\| \bz_{j}^n \|_1} $
\end{itemize}
where $k = argmin_j \sum_{i=1}^m z_{i,j}^o Q_{i,j}^o$, and $\| \bz_{j}^n \|_1 = \sum_{i'=1}^m z_{i'j}^n $
\end{theorem}




\textit{Proof:} In order for an element to become a representative of atleast one incoming frame $i \in X_{t+1}$, it should be having the minimum function value. The first condition states that a new frame $j \in X_{t+1}$ will be selected as a representative in place of $j' \in X_{t+1}$ if it holds a lower function value ($Q^n_{ij} < Q^n_{ij'}$). If $j'$ would have been a representative, $z_{ij'}^n = 1$ would have to be true. This would not be possible due to a higher function value of $Q_{ij}^n$. By contradiction, \textit{Condition 1} holds.

A point $j \in X_{t+1}$ will get selected if $\sum_{i'=1}^m z_{i'j}^n Q_{i'j}^n < \sum_{i'=1}^m z_{i'k}^o Q_{i'j}^o + \lambda \| [z_{1,j}^n \dots z_{m,j}^n ] \|_p $. The second condition states that an element can be a representative from $X_{t+1}$ if its cost is atmost $\lambda$ times higher than the best representative from the existing set $R_t$ (k = $argmin_j \sum_{i=1}^m z_{i,j}^o Q_{i,j}^o$). From the above condition, we can see that $\sum_{i'=1}^m Q^n_{i'j} < \frac{ \sum_{i'=1}^m z_{i',k}^o Q^o_{i'k} + \lambda \| [z_{1,j}^n \dots z_{m,j}^n ] \|_p }{{\| \bz_{j}^n \|_1}} $. This is true for any $i \in X_{t+1}$ which has $j$ as its representative. Hence, we can rewrite the condition as $Q^n_{ij} < \frac{ \sum_{i'=1}^m z_{i',k}^o Q^o_{i'k} + \lambda \| [z_{1,j}^n \dots z_{m,j}^n ] \|_p }{{\| \bz_{j}^n \|_1}} $ where $i \in X_{t+1}$ has $j \in X_{t+1}$ as a representative. Thus, \textit{Condition 2} holds true.




\section{Conditions for Thresholded Convex MCOSS}

\begin{theorem}
The score function in Equation \ref{eqn:newsiftloss} is 
\begin{equation}
\rho (\sum_{i=1}^m \sum_{j=1}^{|R_t|} z_{ij}^{o} d^o_{ij} + \sum_{i,j=1}^m z_{ij}^n d^n_{ij}) - (1 - \rho)( \sum_{j=1}^{|R_t|} S_j^o * L_j^o + \sum_{j=1}^m S_j^n * L_j^n)
\end{equation}

A new point $j$ from the set $X_{t+1}$ is selected as a representative by minimizing the above function, only if all of the following hold:
\begin{itemize}
\item $\sum_{i=1}^m z_{ij} \geq \epsilon$
\item $\rho \sum_{i=1}^m z_{ij}^{n} d^n_{ij}(t) -(1-\rho) (S_j^n * L_j^n) \leq \sum_{i=1}^m z_{ik}^n d^o_{ik}(t) -(1-\rho)(S_k^o * L_k^o) + max(|d_{ik}^o - d_{ij}^n| , |L_k^o - L_j^n|)$
\end{itemize}
where k = $argmin_j (\rho \sum_{i=1}^m z_{ij}^{o} d^o_{ij} - (1- \rho) S_j^o * L_j^o)$
\end{theorem}

\textit{Proof:} The condition for the function value of a representative to be low holds true here similarly. The first condition states that a new frame $j \in X_{t+1}$ is termed as a representative if its total contribution value towards the set of incoming frames is $\geq \epsilon$. The second condition states that an element can be a representative from $X_{t+1}$ if its cost is atmost $max(|d_{ik}^o - d_{ij}^n| , |L_k^o - L_j^n|)$ times higher than the best representative from the existing set $R_t$ (k = $argmin_j \sum_{i=1}^m z_{i,j}^o Q_{i,j}^o$).

\section{Comparison of methods: Relative Angle affordance for 100:7 compression ratio}

We draw a comparison among the subset selection methods on the basis of prediction of the affordance 'Relative Angle' for the compression ratio 100:7.

\begin{figure}[h!]
     \centering
    \begin{subfigure}{0.45\columnwidth}
    \centering
    \includegraphics[width=\textwidth,height=3cm]{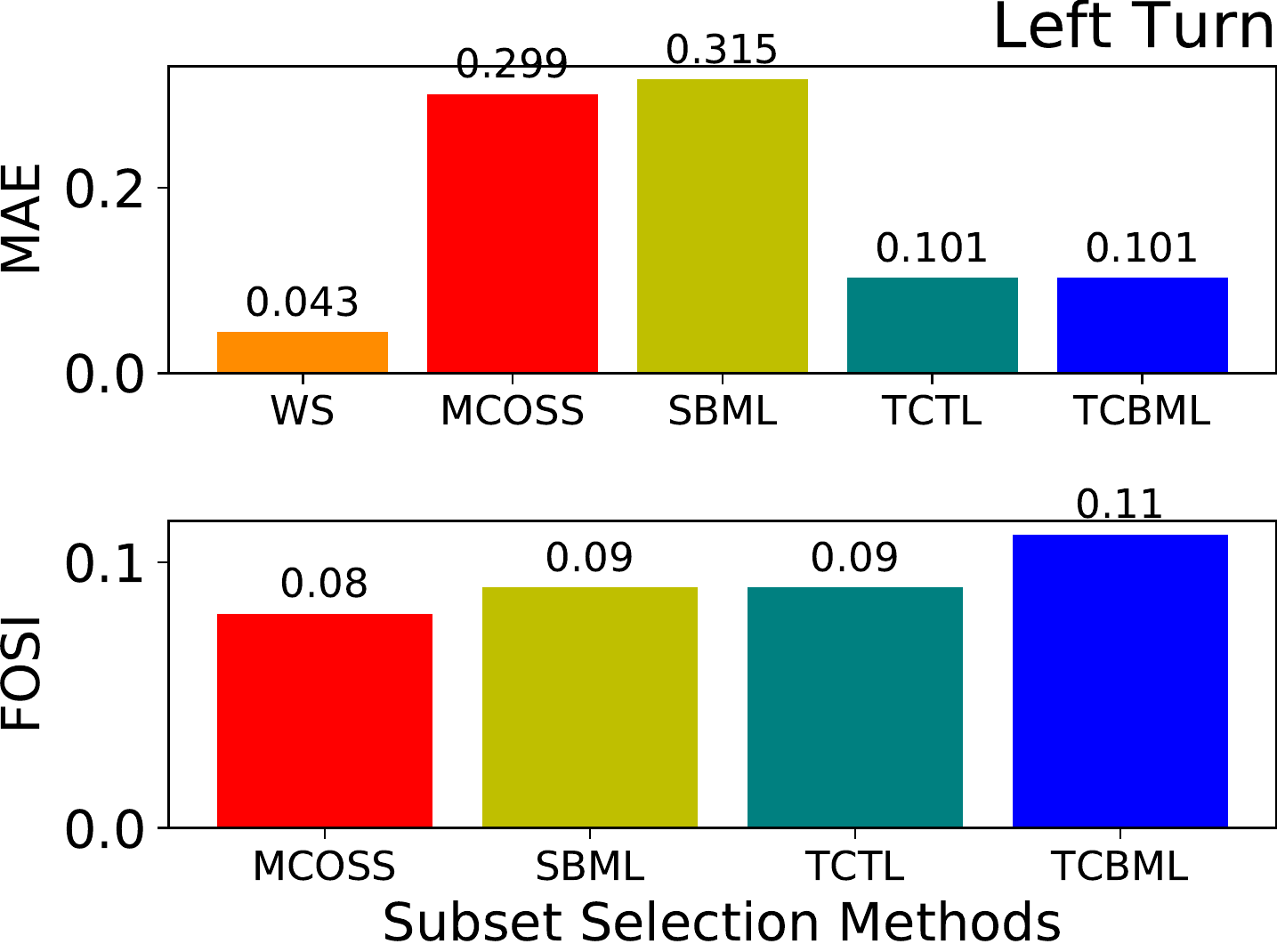}
    \label{fig:tskip15}
    \end{subfigure}
    \centering
    \begin{subfigure}{0.45\columnwidth}
    \centering
    \includegraphics[width=\textwidth,height=3cm]{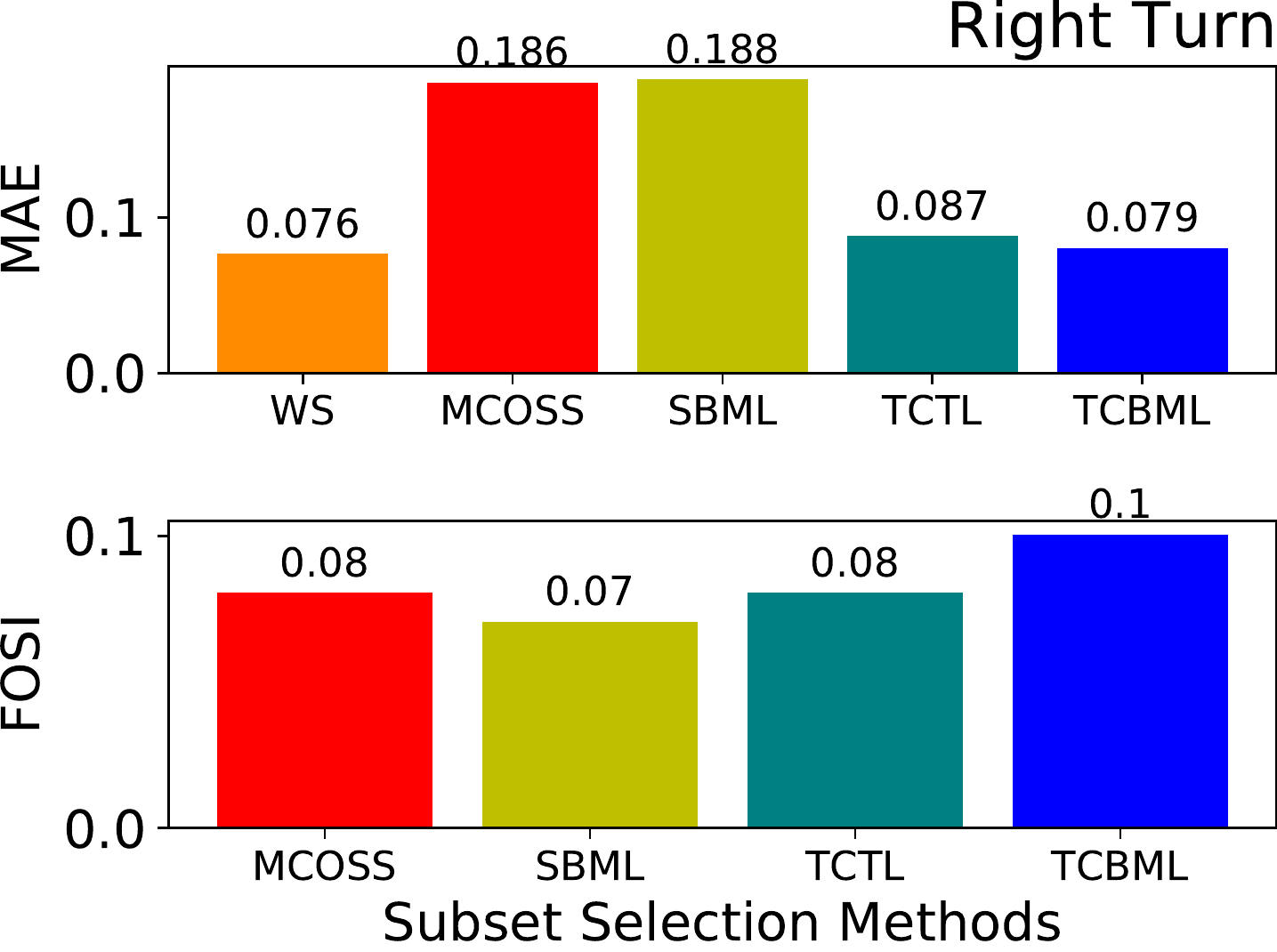}
    \label{fig:tskip5}
    \end{subfigure}
    
    \centering
    \begin{subfigure}{0.45\columnwidth}
    \centering
    \includegraphics[width=\textwidth,height=3cm]{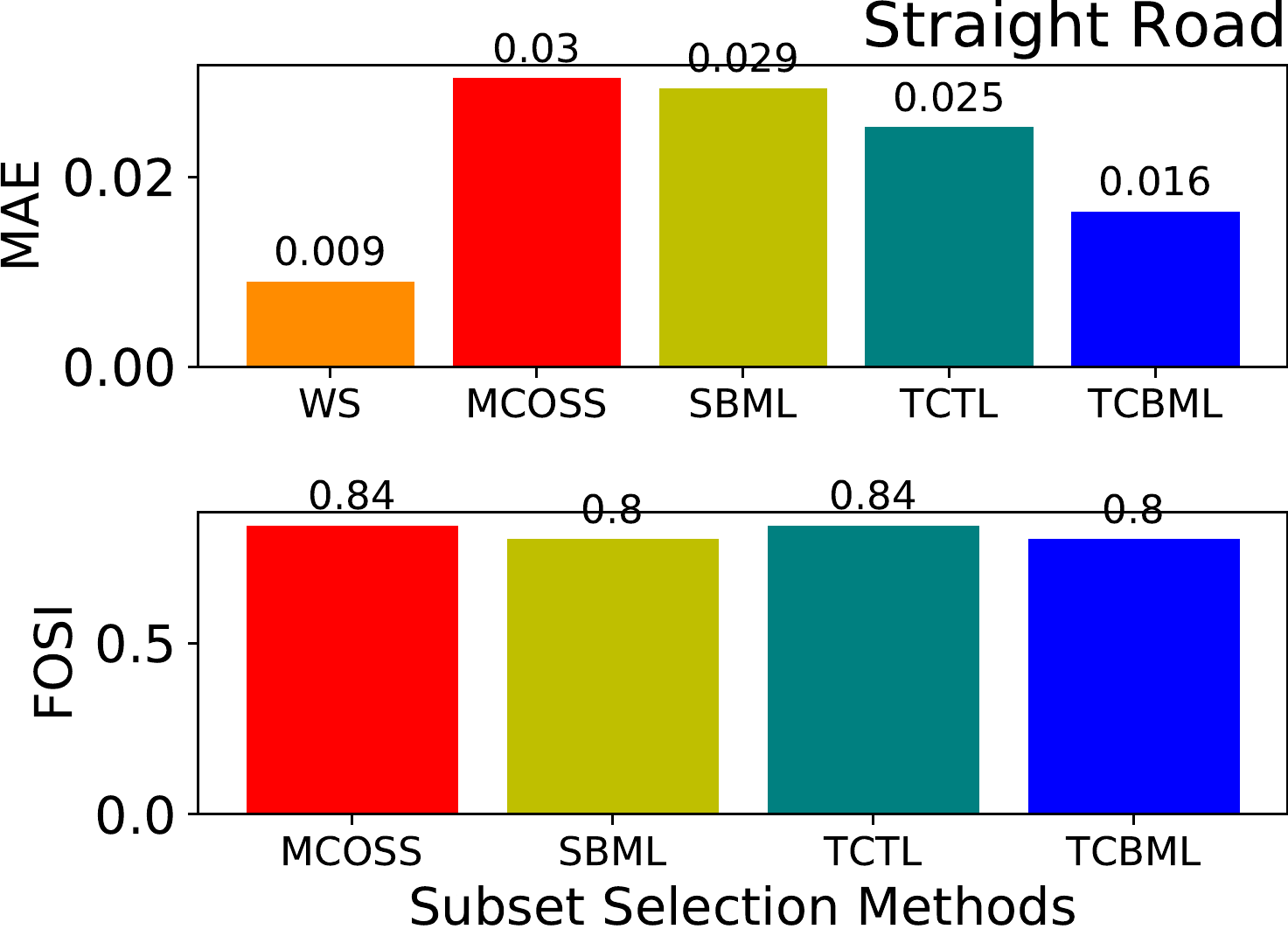}
    \label{fig:tskip5}
    \end{subfigure}
    \caption{\textbf{Fraction of selected instances (FOSI) and Mean Absolute Error (MAE) for Relative Angle affordance for all subset selection methods. Selection of more fraction of instances in turn buckets lead to lower MAE in CBML for 100:7 compression ratio.}}
    \label{fig:fracmae_100_20}
\end{figure}

We can clearly observe that the proposed methods CTL and CBML have a lower Mean Absolute Error (MAE) compared to MCL and SBML. This is essentially due to the effective selection of significant instances which help in better prediction of relative angle values.

\section{Algorithm of the proposed convex formulation}

\begin{align}
\label{eqn:newsiftloss}
    \min_{z^o_{ij},z^n_{ij}} & \rho \sum_{i=1}^m \sum_{j=1}^{|R_t|} z_{ij}^{o} d^o_{ij}(t) + \sum_{i,j=1}^m z_{ij}^n d^n_{ij}(t) \nonumber \\ & -(1-\rho) (\sum_{j=1}^{|R_t|} S_j^o * L_j^o  + \sum_{j=1}^m S_j^n * L_j^n )\\
    &s.t. \sum_{j=1}^{|R_t|} z_{i,j}^o + \sum_{j=1}^m z_{i,j}^n = 1 \nonumber \\
    &z_{i,j}^n, z_{i,j}^o \in [0,1] \nonumber \\ 
    &\sum_{j=1}^{m}\| [z_{1,j}^n \dots z_{m,j}^n ] \|_p \leq frac * m
    \nonumber
\end{align}

Algorithm \ref{algo:thresholdsubset} describes the proposed thresholded convex optimisation method (Equation \ref{eqn:newsiftloss})  for subset selection.
\begin{algorithm}[tb]
 \caption{: Thresholded Convex MCOSS}
 \label{algo:thresholdsubset}
 \begin{algorithmic}[1]
 
 \State \textbf{Input:}
 \State \hspace{2mm}$R_0$: Existing Set of Instances
 \State \hspace{2mm}$X_t$: Incoming Set of Instances at time $t$
 \State \hspace{2mm}$\rho$, \textit{frac}: Set of parameters for the optimization
 \State \hspace{2mm}$M_0$: Trained Model on $R_0$
 \State \hspace{2mm}$T$: Number of incoming batches
 \State \textbf{Process:}
 \hspace{2mm}\For{$t$ = 1,2, $\ldots$, $T$}
 \State  Calculate loss metric $L_j(t), j\in X_t,R_{t-1}$ using $M_{t-1}$
 \State  Calculate SIFT distance $d_{ij}, \forall (i,j)\in X_t\times (R_{t-1} \cup X_{t})$  (refer \cite{lowe1999object})
 \State  Solve optimization problem of Eqn \ref{eqn:newsiftloss} to get $z^o_{ij} \forall (i,j)\in X_t\times R_{t-1}$ and $z^n_{ij} \forall (i,j)\in X_t\times X_{t}$ 
 \State  For each $j\in X_t$ ,calculate its representativeness as $\sum_{i=1}^m z_{ij}^n$. If representativeness $> \epsilon$, add it to $R_{t-1}$.
 \State $R_{t}$ = $R_{t-1}$
 \State Train the model $M_{t}$ using $R_{t}$
 \EndFor
 \State \textbf{Output:}
 \State $M_T$ : Final trained model
 \State $R_T$ : Final reduced dataset
 \end{algorithmic}
\end{algorithm}




{\small
\bibliographystyle{ieee_fullname}
\bibliography{egbib}
}